\documentclass[10pt,twocolumn,letterpaper]{article}

\usepackage{iccv}
\usepackage{times}
\usepackage{epsfig}
\usepackage{graphicx}
\usepackage{amsmath}
\usepackage{amssymb}

\usepackage{booktabs}
\usepackage{bbm}
\usepackage{colortbl}
\usepackage{color,xcolor}
\usepackage{subfloat} 
\usepackage{subcaption}
\usepackage{multirow}
\usepackage{bm}
\usepackage{ifthen}
\usepackage{siunitx}
\usepackage{marvosym}
\usepackage{bbm}
\usepackage{enumitem}
\usepackage{adjustbox}

\usepackage[pagebackref=true,breaklinks=true,letterpaper=true,colorlinks,bookmarks=false]{hyperref}

\iccvfinalcopy 

\definecolor{color3}{rgb}{0.95,0.95,0.95}

\begin{document}

\title{A Preliminary Exploration Towards General Image Restoration}

\author{Xiangtao Kong$^{3}$\thanks{This paper was done while the author was affiliated with Shanghai AI Laboratory and Shenzhen Institutes of Advanced Technology, CAS.} 
\quad Jinjin Gu$^{4}$
\quad Yihao Liu$^{1}$
\quad Wenlong Zhang$^{1,3}$  \\
\quad Xiangyu Chen$^{1,2,5}$
\quad Yu Qiao$^{1,2}$ 
\quad Chao Dong$^{1,2}$ \thanks{Corresponding author (e-mail: chao.dong@siat.ac.cn)}\\
$^{1}$Shanghai AI Laboratory\quad
$^{2}$Shenzhen Institutes of Advanced Technology, Chinese Academy of Sciences\\
$^{3}$The Hong Kong Polytechnic University\quad
$^{4}$The University of Sydney\quad
$^{5}$University of Macau\\
{\tt\small \{xiangtao.kong, wenlong.zhang\}@connect.polyu.hk, jinjin.gu@sydney.edu.au, }
\\
{\tt\small \{liuyihao, qiaoyu\}@pjlab.org.cn, chxy95@gmail.com, chao.dong@siat.ac.cn}
}
\maketitle

\begin{abstract}
Despite the tremendous success of deep models in various individual image restoration tasks, there are at least two major technical challenges preventing these works from being applied to real-world usages: (1) the lack of generalization ability and (2) the complex and unknown degradations in real-world scenarios.
Existing deep models, tailored for specific individual image restoration tasks, often fall short in effectively addressing these challenges.
In this paper, we present a new problem called general image restoration (GIR) which aims to address these challenges within a unified model.
GIR covers most individual image restoration tasks (\eg, image denoising, deblurring, deraining and super-resolution) and their combinations for general purposes.
This paper proceeds to delineate the essential aspects of GIR, including problem definition and the overarching significance of generalization performance. Moreover, the establishment of new datasets and a thorough evaluation framework for GIR models is discussed.
%
%
We conduct a comprehensive evaluation of existing approaches for tackling the GIR challenge, illuminating their strengths and pragmatic challenges.
By analyzing these approaches, we not only underscore the effectiveness of GIR but also highlight the difficulties in its practical implementation.
At last, we also try to understand and interpret these models' behaviors to inspire the future direction.
Our work can open up new valuable research directions and contribute to the research of general vision.
\end{abstract}

\setlength{\abovedisplayskip}{2pt}
\setlength{\belowdisplayskip}{2pt}

\section{Introduction}
\label{sec: Introduction}
%

We present a research problem called General Image Restoration (GIR). 
The primary goal of GIR is to develop a unified and systematic approach to address a wide range of image restoration challenges that may arise in real-world applications. 
The objective of GIR is to transform any degraded input image into a corresponding natural and clear output. 
This encompasses not only individual image restoration tasks such as image denoising, deblurring, and super-resolution, but also the combination of these tasks, as well as real-world image degradations that are currently difficult to model. 
Given the limited existing literature on GIR, we need to construct the entire framework from scratch, including defining the problem, establishing evaluation protocols, developing baseline models, and interpreting these models.
Before proceeding with these steps, it is crucial to specify our motivation for proposing the GIR framework.

Let us begin with the recent progress in image restoration.
%
%
Benefiting from the fast development of deep learning techniques, individual image restoration tasks~\cite{HAT,chen2022simple,TLC,chen2023dual} have enjoyed tremendous success in well-defined settings and constraint environments.
However, they are still faced with the generalization problem and struggle to be applied in real-world usages~\cite{RealESRGAN,BSRGAN,kong2022reflash,liu2022evaluating,gu2023networks}.
Specifically, real-world images frequently present complex and unpredictable degradations, overwhelming most existing restoration models.
For example, a hazy image might also exhibit defocus blur and JPEG compression, rendering a task-specific dehazing model insufficient for such complex cases.
Furthermore, even if the hazy image does not contain other degradations, its haze feature might deviate from the training data distribution, leading to suboptimal results.
The challenge of generalization is the primary driving force behind our proposal of the GIR problem. 
There is a pressing need for a comprehensive model capable of addressing diverse real-world challenges without necessitating specific parameter adjustments.

\begin{figure*}
    \centering
    \includegraphics[width=0.7\linewidth]{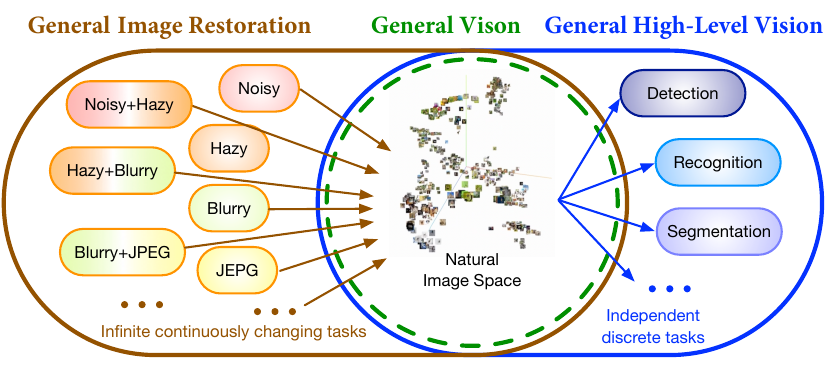}
    \vspace{-2mm}
    \caption{The differences between the General Image Restoration (GIR) and the General High-level Vision (GHV).}
    \label{fig:GIR_def}
    \vspace{-2mm}
\end{figure*}

From another aspect, we can observe a similar trend in high-level vision.
While conventional high-level vision tasks (\eg, image classification and object detection) have approached their performance bottleneck, general vision models (GVMs) came into view and developed at an astonishing speed.
For example, INTERN \cite{shao2021intern} could deal with tens of vision tasks and achieve state-of-the-art performance with a single general model.
Visual models based on multimodal language models can also complete a variety of visual tasks and have considerable intelligence.
It is well believed that GVM is a promising direction to solve the generalization problem.
Nevertheless, existing GVM experience and techniques are not suitable for image restoration.
%
%
First, while other models take high-dimensional image data as input and obtain low-dimensional labels, image restoration models have both the input and output in the form of images. The generation of images is the problem that GIR needs to consider. 
Secondly, GVM extracts semantic information hierarchically from clear images, whereas image restoration models not only need to capture abstract semantic information but also handle pixel-level information in degraded images.
More importantly, GIR is NOT a naive extension of multi-task problem!
It should deal with real unseen data and complex degradation combinations, as depicted in \figurename~\ref{fig:GIR_def}.
Recently, there are also some low-level vision models similar to the concept of GVM have been proposed\cite{GenLV,PromptGIP}. %
Instead of focusing on restoration, they focus more on processing images from one modality to another.


With the above background, we can describe the necessities of studying GIR.
First and foremost, GIR is a touchstone of generalization ability.
As deep models are bone to overfit the training data, it is impossible to solve the generalization problem with a single-task model.
GIR covers not only individual image restoration tasks but also their combined tasks (\eg, deraining $+$ denoising).
Thus, it naturally favours the model with better generalization.
On the other hand, when combining various tasks, we hope ideal deep models to learn the natural image distribution instead of specific degradations, providing a feasible solution towards a real general model.
This also meets GIR's requirements for intelligence. 
Except for the research value, GIR also has great application demand.
Note that real-world scenarios usually cannot be synthesized by mathematical models.
Complex and unknown degradations can appear in various forms, such as underwater blur and old film corruption, which cannot be processed by dedicated deep models.
Therefore, we expect a general model that is powerful enough to handle real complex cases.
Furthermore, it is also desirable for casual users to have a general model in hand, which can process most images without model selection.

To this end, we propose the new problem GIR and start with the most fundamental definition.
At the beginning, we will clarify the differences between general high-level vision models, multi-task and blind image restoration models.
After a detailed comparison, we can figure out the uniqueness of GIR.
Then we will build the evaluation protocol, including degradation models, test datasets and evaluation metrics.
This part mainly tells us ``what tasks should be covered'', ``where to evaluate the image quality'' and ``how to evaluate the generalization performance''.
With such an evaluation platform, we can benchmark existing potential methods.
After careful investigation, we show the significance of GIR but find that most deep learning architectures and training strategies are not feasible for the GIR problem.
It is also surprising that Transformers do not always outperform CNNs under such a complex setting.
We also provide insightful discoveries from image restoration interpretation methods, such as LAM \cite{gu2021interpreting} and DDR \cite{liu2021discovering}.
These facts inspire us that we need completely different networks and training strategies to realize GIR, as well as to solve the generalization problem.
We hope our work can lay the foundation towards a real general model in image restoration. 

Our contributions can be summarized as follows:

\begin{enumerate}
    \item We present a new GIR problem, which aims at processing unnatural and degraded images into natural and clean ones with one model. 
We discuss the necessity of GIR for real-world applications, and the uniqueness of GIR from existing technologies.
    \item We provide a preliminary research framework for the GIR problem, including a simplified definition, dataset construction and evaluation methods.
Our research framework allows us using existing techniques to explore GIR problems.
    \item We build a benchmark for the existing possible GIR approaches. We verify the necessity, effectiveness and interpret the difficulties of GIR. Our results shed light on the future research of general vision.
\end{enumerate}




\section{Related Work}
\paragraph{Generalization Problem in Image Restoration.}
Researchers have achieved great success in various individual image restoration tasks with mutually matched training and test sets, such as image denoising, super-resolution, deblurring, deraining, \etc~\cite{SRCNN,SRResNet,EDSR,NLSA,PAN,zhang2017beyond,SRN,FFDNet,db,dong2015compression,Diffjpeg,pathak2016context,Dehazenet,yang2017deep,RESIDE,PReNet,derainnet,GridDehazeNet,PFDN,DehazeFormer,Rain100,zhang2018density,desnownet,chen2021all,HINet,MPRNet}.
However, when the test degradation does not match the training degradation, many image restoration models will fail, \eg, different downsampling kernels in super-resolution \cite{gu2019blind,liu2022blind,kong2022reflash,zhang2023crafting}, different noise distributions \cite{SIDD,chen2023masked} and different rain streaks \cite{Rain100,gu2023networks}.
This lack of generalization ability poses a serious obstacle to the application of image restoration models, because the degradation of real-world input images is often different from the training set.
Existing works either develop blind models to include more degradation possibilities during training, or bring the training data closer to practical applications.
Existing blind restoration models often assume a pre-defined degradation model and assume that some of its parameters are unknown.
However, the coverage of these degradation models is also very limited.
For example, many blind super-resolution works assume that the blur kernel is Gaussian \cite{gu2019blind,liu2022blind}.
For situations beyond this range, these methods still face generalization problems.
Another approach is to collect training images as similar to the real situation as possible \cite{SIDD,cai2019toward}.
But this approach can only cover a very limited range of applications and is costly.
Collecting training images for a broad range of applications is impractical.
Only a little work has been proposed to study the reasons for the lack of generalization ability in image restoration.
Kong \etal \cite{kong2022reflash} propose to improve the generalization performance of SR networks by dropout layers.
Liu \etal \cite{liu2021discovering} present that networks tend to overfit to degradations and show deep degradation ``semantics'' inside the network.
The existence of these deep degradation representations often means a decline in generalization ability.
The utilization of this knowledge can guide us to analyze and evaluate the generalization performance \cite{liu2022evaluating}.
Gu \etal \cite{gu2023networks} and Chen \etal \cite{chen2023masked} try different training strategies to understand the generalization of the derain and denoise task.
Besides that, few works have been proposed to improve the generalization ability of models.
Regarding cross-task generalization (specifically, handling degradation types not seen during training) there is no existing work.

\paragraph{Multi-Task Image Restoration.}
Multi-task image processing methods have been extensively studied.
Three classes of work are closely related to the discussion in this paper.
First, multi-task pre-training-based methods represented by IPT \cite{IPT}, EDT \cite{EDT} and DegAE~\cite{DegAE} are closely related to GIR.
These methods utilize a shared backbone network for different tasks and replace different processing heads and tails for each task.
Secondly, some methods~\cite{Chen2022MultiWeatherRemoval,zhang2023ingredient,liu2022tape,ProRes,PromptIR,li2023prompt,kong2024towards,xrestormer,GenLV,PromptGIP} use a unified structure to handle multiple tasks.
Although they can handle multiple image processing tasks with one model, they only focus on specific tasks, and need to know degradation types. 
Thus, these methods cannot solve the generalization problem, nor can they handle cross-task or mixed situations.
Another class of methods utilizes a network to learn tasks with complex degradations, such as Real-ESRGAN \cite{RealESRGAN} and BSRGAN \cite{BSRGAN}.
These methods consider the limited crossover between tasks and enable the network to generalise on these tasks.
But for the cross-degradation of multiple broader tasks (\eg noise + rain), we have yet to witness related work.
Besides that, their evaluation systems are not clear and inadequate enough.
%


\paragraph{General Models and General High-level Vision.}
The development trend of artificial intelligence models is undergoing a paradigm shift from dedicated task-specific to general.
First, general language models (GLM) have achieved impressive progress.
Through large-scale pre-training, many models such as BERT \cite{BERT} and GPT-3 \cite{GPT3}, have demonstrated the potential of general models.
These models greatly benefit a wide range of language-related downstream tasks through task-specific fine-tuning and adaptation.
Furthermore, with the task-independent training objectives and pretext tasks \cite{MAE}, the performance gain of pre-training can be increased by employing large-scale unlabeled data and a very large model.
The success of GLM has also inspired the General Vision Model (GVM).
GVM attempts to learn general representations in the computer vision field and use one general model to cover a wide range of vision tasks.
Many previous works employ large-scale supervised \cite{zhai2022scaling,riquelme2021scaling,dai2021coatnet,sun2017revisiting,deng2009imagenet}, self-supervised \cite{chen2021exploring,grill2020bootstrap,chen2020simple,chen2021empirical,bao2021beit,caron2021emerging}, and cross-modal \cite{huo2021wenlan,radford2021learning} pre-training and show generality on a limited scope of downstream vision tasks.
Among these works, ViT-G \cite{ViT-G} uses categorical supervision, SEER \cite{SEER} applies contrastive learning between different augmented views, while CLIP \cite{CLIP} uses paired language descriptions, and INTERN \cite{shao2021intern} first learns expert models on different tasks and then integrates them into a general model.
The main text discusses the difference between general image restoration and general high-level vision.
These differences make it difficult to directly apply the experience accumulated in general high-level vision to the development of general image restoration.
This paper has illustrated this from the perspective of definition, task construction and model analysis.



\section{General Image Restoration}
\label{sec: General Image Restoration}
\subsection{Definition}
\label{sec: Definition}
\begin{figure}[t]
  \centering
  \includegraphics[width=\linewidth]{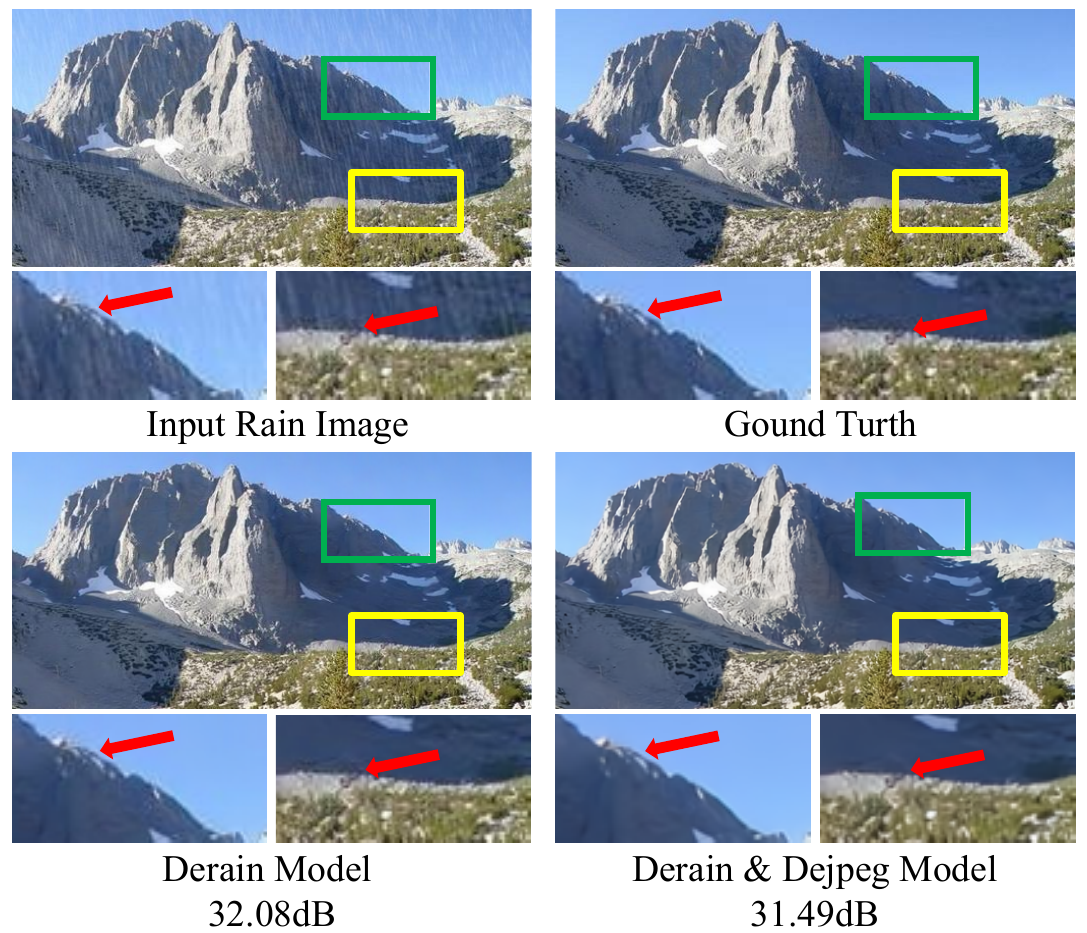}
  \caption{The Rain1200 result of models with different ability. The resulting image of derain\&djpeg model obtains lower PSNR value, even though it looks more natural and cleaner. This issue arises from the reference GT images lacking the desired level of pristine quality and idealized perfection.}
  \label{fig:derain}
  \vskip -0.2cm
\end{figure}

%
As general image restoration (GIR) is a new topic, we need to give it a primary and reasonable definition.
Broadly speaking, we expect GIR to cover all image restoration tasks.
However, many existing individual tasks even have different goals.
For example, deraining task only focuses on the effect of rain removal.
Hence even almost no-rain images of deraining test sets are suffered from compression artifact~\cite{Rain100,zhang2018density}, it is acceptable.
But it is inapposite for GIR. (We discuss this state in Section\ref{sec:testset} and Figure ~\ref{fig:derain})
GIR is not a simple mix of existing tasks. It needs to consider mixed degradation and focus on the common goal of restoration.
Therefore, we first assume a desired image space (\eg, natural and clear).
The images in this space will be considered in line with human subjective aesthetic judgments.
A GIR model can recover ``commonly observed'' degraded images to the corresponding desired natural and clean states, rather than focusing on a few specific types and ignoring the rest.
To make this definition more concrete, we compare it to general high-level vision (GHV), multi-task image restoration (MIR), and blind image restoration (BIR).


\paragraph{GIR \vs GHV.}
Interestingly, the pipelines of GIR and GHV are somewhat symmetrical, as shown in \figurename~\ref{fig:GIR_def}.
Ideally, the output space of GIR and the input space of GHV are the same natural image space.
By combining GIR and GHV, we can naturally work toward a general vision model.
Moreover, the difference between GIR and GHV also lies in their evaluation manner.
GHV has clear human labels for accurate counting, but GIR needs to evaluate the image quality, which is subjective and ambiguous.

\paragraph{GIR \vs MIR.}
MIR is the closest concept to GIR in literature, but is different from GIR in two main aspects.
First, they have different input image/task spaces.
The input space for MIR is discrete (several pre-defined degradation/tasks), but for GIR it is continuous. 
MIR contains multiple disjoint restoration tasks, but GIR also includes their cross operations (e.g., denoise mixing deblur). 
Compared to the limited, well-defined tasks in MIR, the number of possible degradations in GIR is very large, as the combinations of degradations are nearly infinite. 
Second, the input information for MIR and GIR is also different. 
The model needs to know all the input degradation types beforehand for MIR. 
However, for GIR, the degradation information is unnecessary.
The two differences above also apply to single-task restoration.
In a word, GIR is a ``blind'' problem that covers a wide range of degradations and their combinations. 
It is impractical to assign task-specific parameters to different tasks. 

\paragraph{GIR \vs BIR.}
As mentioned above, GIR is a ``blind'' problem.
However, the most existing BIR methods are developed for a specific task, \eg, blind SR and blind deblurring.
They usually adopt a known degradation model but treat the degradation parameters as to-be-predict factors.
These methods also tend to take advantage of pre-defined degradation models as priors.
But for GIR, we can not define such a parametric model. 
Most of the BIR methods (\eg, DAN \cite{luo2020unfolding,luo2021endtoend} and IKC~\cite{IKC}) cannot be directly applied to GIR.
GIR should have an extensible pipeline to adopt any type of newly appeared degradations and have an clear evaluation system to evaluate generalization ability.

\subsection{Problem objective}
\label{sec:objective}
%
To better distinguish the objective of GIR from existing research objectives, we start with the objective of MIR problem.
We represent the MIR input space as $\{\mathcal{X}^t\}_{t\in[T]}$, and the target image space as $\mathcal{Y}$ where $T$ is the number of tasks.
The dataset is collected as $\{\mathbf{x}_i^1,\dots,\mathbf{x}_i^T,\mathbf{y}_i\}_{i\in[N]}$, where $N$ is the number of data points, $\mathbf{x_i}$ and $\mathbf{y_i}$ are input and ground truth images.
Similar to multi-task learning \cite{sener2018multi}, the model is given for each task by $f^t(\mathbf{x}^t;\theta^{\mathrm{sh}},\theta^t):\mathcal{X}^t\mapsto\mathcal{Y}$, where $\theta^{\mathrm{sh}}$ is the model parameters shared between tasks and $\theta^t$ is the task-specific parameters.
The loss function for each task is given by $\mathcal{L}^t(\cdot,\cdot):\mathcal{Y}\times\mathcal{Y}\mapsto\mathbb{R}^{+}$.
The current MIR paradigms generally yield the empirical risk minimization objective:
%
%
\begin{equation}
    \min_{\theta^{\mathrm{sh}},\theta^1,\dots,\theta^T}\sum_{t=1}^T w^t\hat{\mathcal{L}}^t(\theta^{\mathrm{sh}},\theta^t),
    \label{eq:MIR}
\end{equation}
where $\hat{\mathcal{L}}^t(\theta^{\mathrm{sh}},\theta^t):=\frac{1}{N}\sum_{i=1}^N\mathcal{L}^t(f^t(\mathbf{x}^t;\theta^{\mathrm{sh}},\theta^t),\mathbf{y}_i)$.
This is an intuitive practice of MIR that can be implemented using a simple multi-objective weighted optimization method given the type of task. 
Accordingly, MIR evaluation only focuses on these tasks.

Compared to the MIR problem, the objective of GIR differs in two aspects. 
First, compared to the limited, well-defined tasks in MIR, the number of possible degradations in GIR is very large, as the possibilities of degradation combinations are nearly infinite. This implies that in GIR, the number of tasks $T^\mathrm{GIR}\to\infty$.
Second, due to the large number of potential tasks, it is impractical to assign task-specific parameters for each task.
Thus the model parameters are shared among all tasks as $\theta^\mathrm{GIR}$.
Let $\{\mathcal{X}^t\}_{t\in[T^\mathrm{GIR}]}$ be the set of all possible tasks involved in the GIR problem.
The objective formulation at this point can be written as
\begin{align}
    \min_{\theta^{\mathrm{GIR}}}\sum_{t=1}^{T^\mathrm{GIR}} w^t\hat{\mathcal{L}}^t(\theta^\mathrm{GIR})+\mathcal{G},
    \label{eq:GIR1}
\end{align}
where $\mathcal{G}:=g(\hat{\mathcal{L}}^1(\theta^\mathrm{GIR}),\dots,\hat{\mathcal{L}}^{T^\mathrm{GIR}}(\theta^\mathrm{GIR}))$ is a functional definition on the model space to measure the model's generality.
This generality term $\mathcal{G}$ means that a GIR model should have acceptable performance on all covered tasks, rather than performing well on some tasks but failing on the others.
However, in practice, we can only access a very limited number of tasks during optimization, \eg, $T^+\ll T^\mathrm{GIR}$ accessible tasks.
The GIR problem yields the objective of
\begin{align}
    \min_{\theta^{\mathrm{GIR}}}\sum_{t=1}^{T^+} w^t\hat{\mathcal{L}}^t(\theta^\mathrm{GIR})+\mathcal{G}+\mathcal{R},
    \label{eq:GIR-opt}
\end{align}
where $\mathcal{R}$ is the risk term of the model on tasks that cannot be sampled.
$\mathcal{R}$ is also a functional defined on the model space to represent the generalization ability of the model and may take many broad forms.
Equation~\eqref{eq:GIR-opt} exploits a similar idea of structural risk minimization to represent the objective of the GIR problem, that is, GIR should focus on the generalization ability and generality of the model.
\begin{figure*}[t]

\scriptsize
\centering
\resizebox{1\textwidth}{!}{
\begin{tabular}{cc}
\begin{adjustbox}{valign=t}
    \large
    \begin{tabular}{ccccc}
    \includegraphics[width=44mm, height= 29mm]{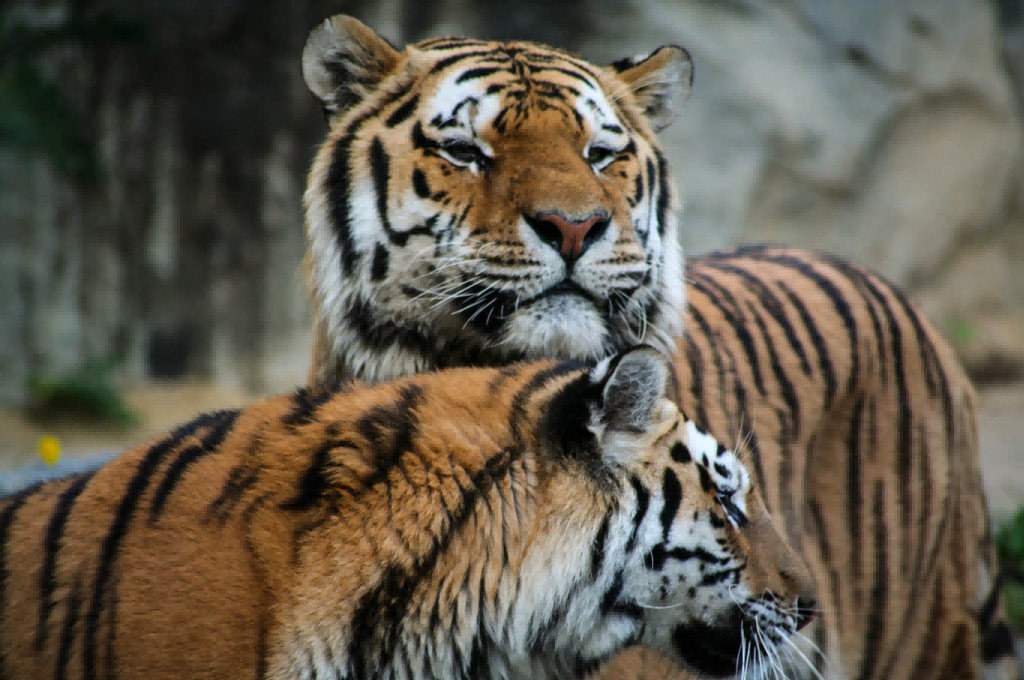}  &
    \includegraphics[width=44mm, height= 29mm]{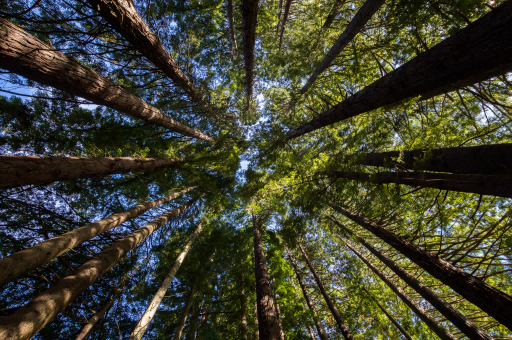}  &
    \includegraphics[width=44mm, height= 29mm]{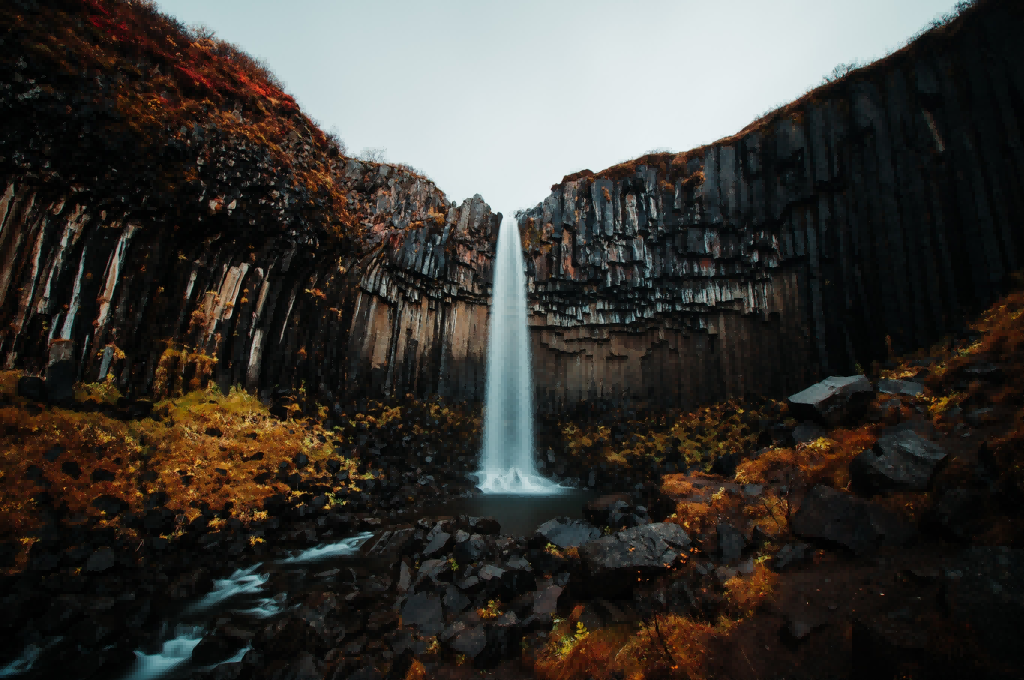}  &
    \includegraphics[width=44mm, height= 29mm]{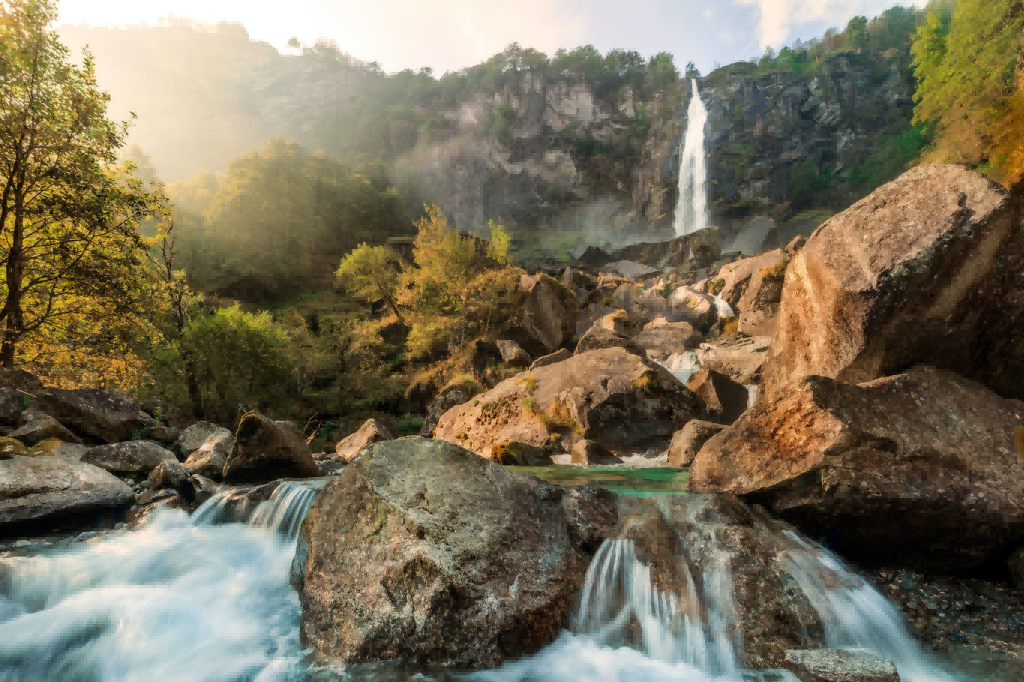}  &
    \includegraphics[width=44mm, height= 29mm]{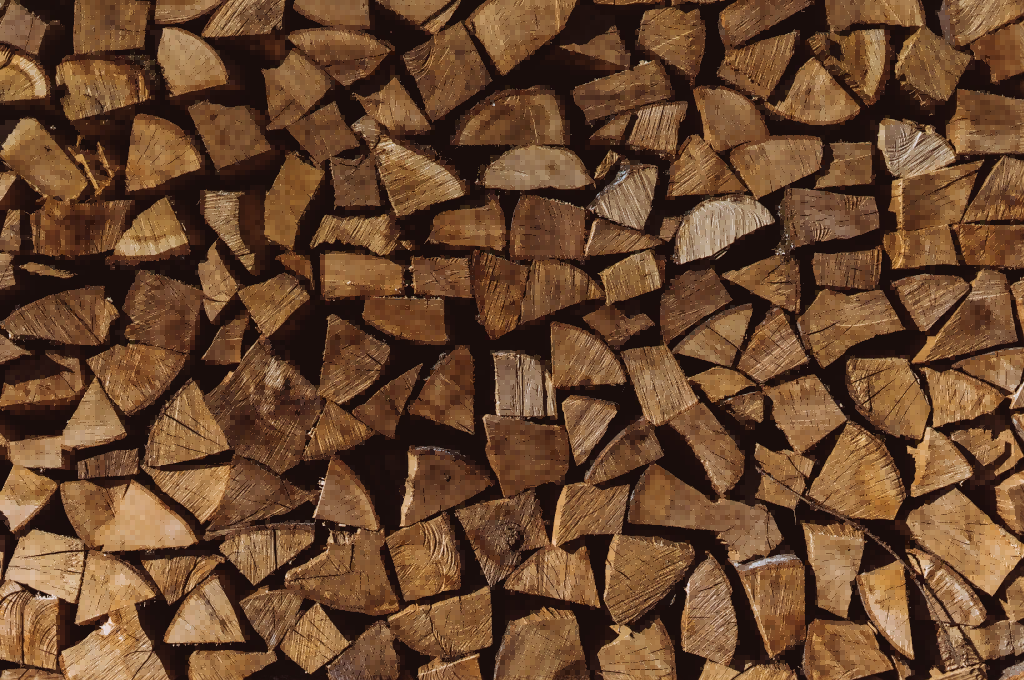} 
    \vspace{1mm}
    \\
    
    Animal  &
    Vegetation  &
    Mountain View &
    Scenery  &
    Texture
    \vspace{1mm}
    
    \\
    \end{tabular}
    \end{adjustbox}
\\
\begin{adjustbox}{valign=t}
    \large
    \begin{tabular}{ccccc}
    \includegraphics[width=44mm, height= 29mm]{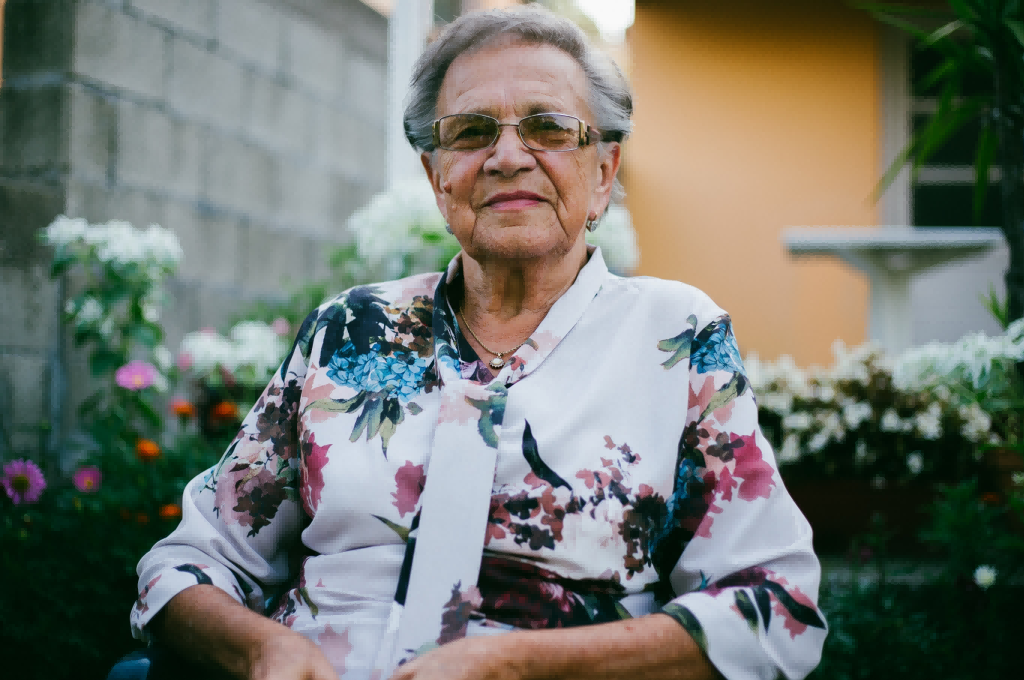}  &
    \includegraphics[width=44mm, height= 29mm]{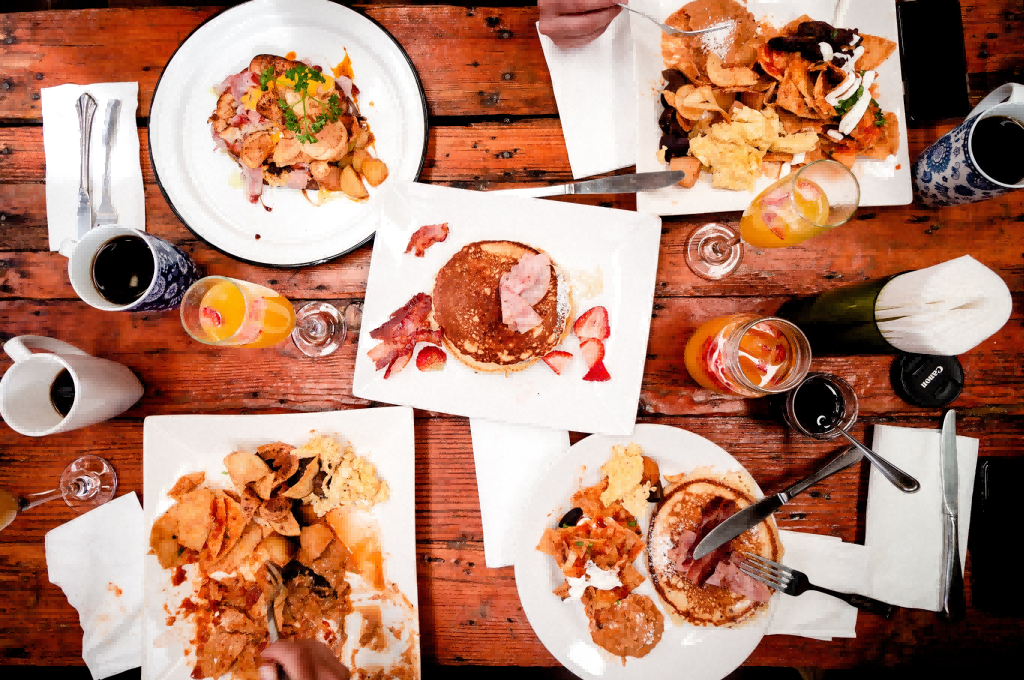}  &
    \includegraphics[width=44mm, height= 29mm]{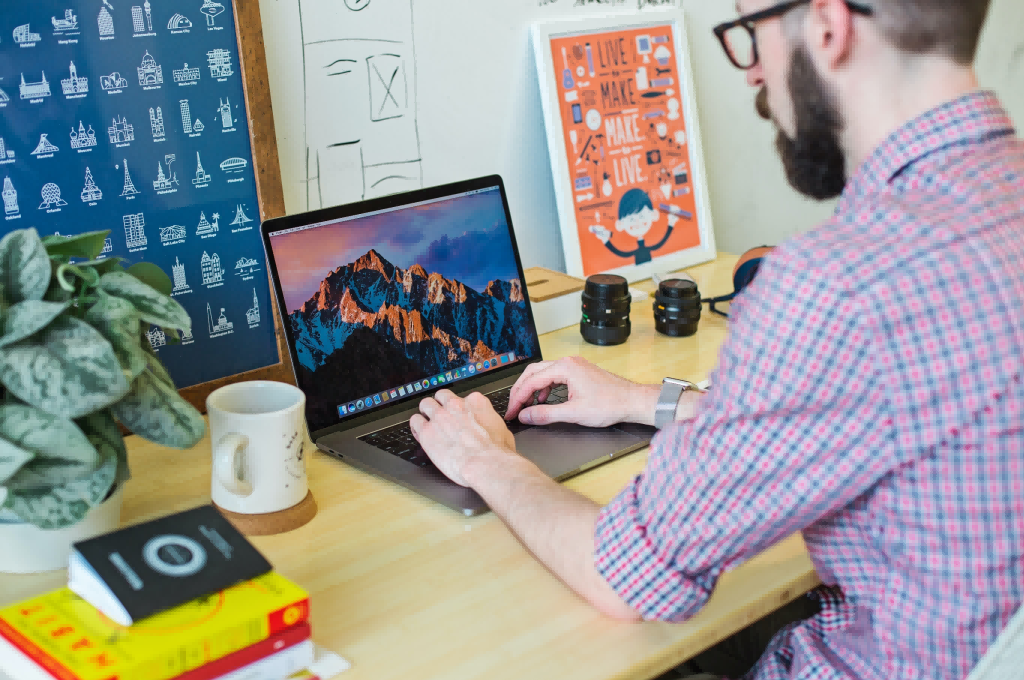}  &
    \includegraphics[width=44mm, height= 29mm]{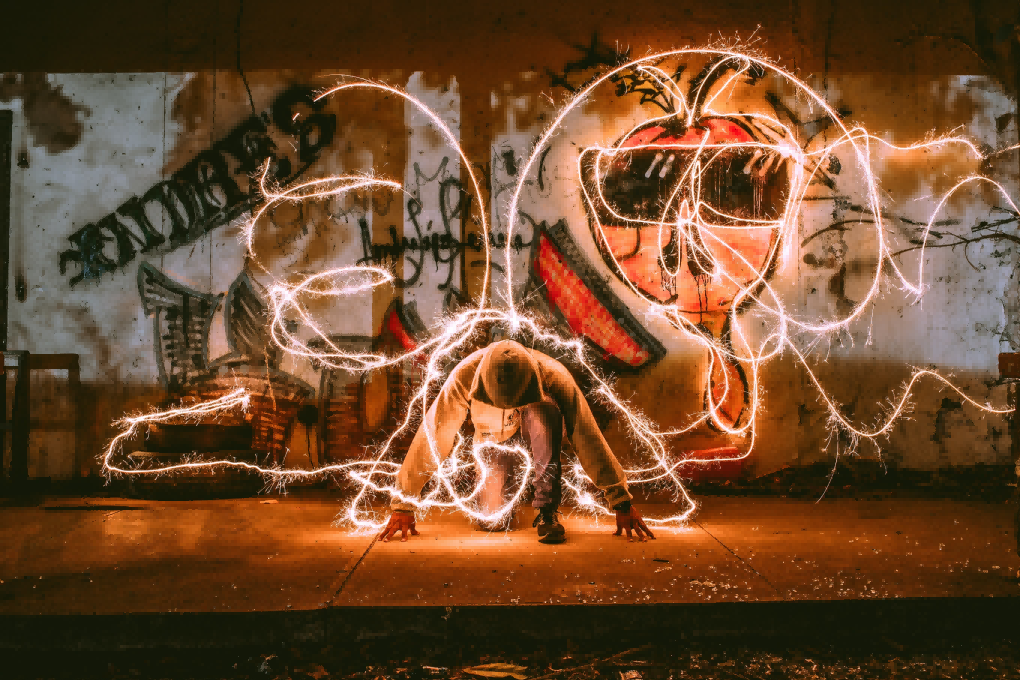}  &
    \includegraphics[width=44mm, height= 29mm]{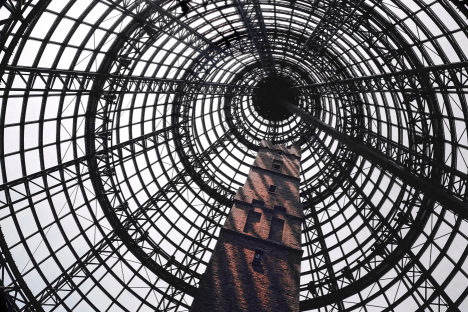} 
    \vspace{1mm}
    \\
    
    Portrait  &
    Food  &
    Daily Scenes &
    Art  &
    Architecture
    \vspace{1mm}

    \\
    \end{tabular}
    \end{adjustbox}
\end{tabular}}
\vskip -0.2cm
\caption{The demo of ground truth images. They contain 10 scenes with 10 images per scene from Unsplash~\cite{unsplash}.}
\label{fig:gt}
\vskip -0.23cm
\end{figure*}

Equation~\eqref{eq:GIR-opt}  not only provides a reference for us to develop GIR methods, but most importantly, it points out three important aspects of our GIR evaluation.
First, we need to evaluate the effect of the GIR model in the tasks covered by the training set.
It is the most basic aspect and is reflected in the first term.
Second, we need to construct a suitable $\mathcal{G}$ to evaluate the generality of the GIR model.
Previous methods that simply compute the average across all tasks cannot objectively illustrate the generality of the model.
Finally, we want to evaluate the performance of the GIR model outside the training set, which is what $\mathcal{R}$ represents.
There are many possible ways to integrate $\mathcal{G}$ and $\mathcal{R}$ into optimization. In this work, we mainly discuss their form in evaluation. We hope to first establish a research framework to facilitate subsequent exploration.

\section{Evaluation for General Image Restoration}
\label{sec: Evaluation for General Image Restoration}
In this section, we describe the evaluation protocol for the GIR approach.
To evaluate the performance across tasks, we first contribute a test set containing selected tasks and testing images in Section~\ref{sec:testset}.
According to the problem objective in Section~\ref{sec:objective}, we propose an effective approach to evaluate the generality $\mathcal{G}$ in Section~\ref{sec:generality} and a feasible way for $\mathcal{R}$ using the real-world data in Section~\ref{Real-world}.

\subsection{Building GIR Test Set}
\label{sec:testset}
Our test set contains both synthetic and real images.
We use MiO100~\cite{kong2024towards} test set which contains 100 high-quality images including 10 scenes and select a series of different image degradations to construct 100 synthetic test sets (10,000 images in total).
For the real images, we also collect 10 types of real-world testing images (each type contains 100 images, 1,000 in total).


\paragraph{Ground truth images.}
Many MIR works~\cite{AirNet,zhang2023ingredient,liu2022tape} directly use synthetic test sets (e.g., Rain1200~\cite{rain1200} and RESIDE~\cite{RESIDE}) from each original single task. 
As mentioned in Section~\ref{sec: Definition}, we think these test sets are inappropriate for GIR and partial MIR tasks.
Because the quality of the ground truth (GT) images can affect their evaluation.
For example, as shown in Figure~\ref{fig:derain}, the GT image from Rain1200 test set contains obvious JPEG compression artifacts.
We can discover that the derain model removes rain but retains JPEG artifacts, while another derain\&dejpeg model can remove rain and JPEG artifacts at the same time.
However, even though the derain\&dejpeg model obtained the result with better quality, it gets a lower PSNR value than the derain model result.
This is because the results of the derain model are more similar to the GT, while the results of the derain\&dejpeg model are more different from GT. 
From this case, it can be seen that lots of so-called ``GT'' images in previous datasets are not actually ``GT''. 
They suffer from various distortions, deviating from ideal natural and clean images..
Note that, these test sets are acceptable for their original single tasks, since these single tasks only focus on their own degradations.
However, they are inappropriate for MIR or GIR problems.

For GIR, the ground truth images should be the desired image state -- high-quality natural images.
Therefore, for synthetic images, we need high-quality images without any distortion and then synthesize low-quality images.
%
%
Besides that, the GIR model is expected to perform well in a variety of different image content scenarios.
However, the images in existing test sets often fail to cover a wide variety of scenes simultaneously.
For these two reasons, we use MiO100~\cite{kong2024towards} test set which contains 100 high-quality images in a variety of different image content scenarios.
These images contain 10 scenes with 10 images per scene from Unsplash~\cite{unsplash}.
As shown in Figure~\ref{fig:gt}, these scenes encompass \texttt{animal}, \texttt{vegetation}, \texttt{mountain view}, \texttt{scenery}, \texttt{texture}, \texttt{portrait}, \texttt{food}, \texttt{daily scenes}, \texttt{art} and \texttt{architecture}.
%

%
%

\paragraph{Degradations.}

Recall that GIR aims to convert 'commonly observed' degraded images into their corresponding desired states, including degraded images from cross-operation between different degradations.
We select a series of different image degradations to construct the test set.
We hope that these tasks can: (1) cover most of the existing well-defined image restoration tasks; (2) focus on reasonable combinations and the mixture of them; (3) be representative for the feasibility of testing.
Following these principles, we introduce the following designs.

1) \emph{For the basic single tasks}, as shown in \tablename~\ref{table:downsampl}, we select 10 existing well-defined tasks according to their commonality.
First, some of them are image degradations caused by digital image acquisition, sensor and storage limitations, \eg, \texttt{resize}, \texttt{blur} and \texttt{noise}.
These are the most common factors for image quality degradation and are almost inevitable.
Second, we also consider the quality degradation of images under artificial post-processing, such as \texttt{compression}, \texttt{ringing} artifacts after image enhancement, and artifacts generated by restoration algorithms \texttt{alg.artifact}.
Finally, we also include bad weather situations such as \texttt{rain}, \texttt{haze} and \texttt{snow}.
This can help machine vision systems combat the degradation of the visual signal caused by external situations.
A typical example is the vision system robustness of an autonomous driving system in bad weather.

%
These degradations are well modeled and there are many solutions available for reference, which is beneficial for building new tasks.
On the other hand, the specific selection of ten tasks is not the key point, the point is that the degradations are already complex enough for helping existing methods to approach GIR. 
A GIR model should have the ability to solve these situations at least.

\begin{table}[t]
    \centering
    \scalebox{0.76}{
    \begin{tabular}{cll}
    \toprule
        \rowcolor{color3} ID & Task Name & Scenario \\
         \midrule
         (1) & \texttt{resize} & low-resolution.  \\
         (2) & \texttt{blur} & \eg, motion blur, unfocus blur. \\
         (3) & \texttt{noise} & \eg, ISP noise.  \\
         (4) & \texttt{compression} & \eg, JPEG artifacts.  \\
         (5) & \texttt{ringing} & ringing/over-sharpening artifacts.  \\
         (6) & \texttt{alg.artifact} & artifacts generated by restoration algorithms. \\
         (7) & \texttt{damage} & parts of pixels are missing.\\
         (8) & \texttt{rain} & visual information is obscured by rain.\\
         (9) & \texttt{snow} & visual information is obscured by snow.\\
         (10) & \texttt{haze} & visual information is obscured by haze.\\
    \bottomrule
    \end{tabular}}
    \caption{The list of the selected 10 basic single tasks.}
    \label{table:downsampl}
    \vspace{-6mm}
\end{table}


2) \emph{For the combination and mixture tasks}, we propose a pipeline to synthesize various mixed degradation scenarios that may occur in the real world.
Inspired by recent works~\cite{RealESRGAN,BSRGAN}, we treat the degradation mixture as a composition of multiple basic degradation functions:
\begin{equation}
  \mathbf{x}= (\mathcal{D}_k \circ \cdots \circ \mathcal{D}_2 \circ \mathcal{D}_1) ( \mathbf{y}), 
  \label{eq:degradations}
\end{equation}
where $\mathcal{D}_i$ represents the degradation function mentioned above with different hyper-parameters, $k$ is the order number of degradation process.
Eq.~\eqref{eq:degradations} is intuitive. The quality degradation of an image may be the result of multiple operations.
For example, taking a photo on a hazy day, the image is enhanced by algorithm processing, resulting in the \texttt{ringing} artifacts, and then compressed during network transmission.
Such a complicated process cannot be simply modeled with single pre-defined degradation models.
In fact, images may go through a random number of random degradation processes.
In our pipeline, we introduce this stochastic mechanism to cover the possible cases as much as possible.
There is only one known exception.
The degradations of \texttt{rain}, \texttt{haze} and \texttt{snow} cannot be added afterwards, but only appear in the process of image acquisition.
Therefore, they can only appear as the first degradation process $\mathcal{D}_1$. 
%

\begin{figure*}[t]

\scriptsize
\centering
\resizebox{1\textwidth}{!}{
\begin{tabular}{cc}
\begin{adjustbox}{valign=t}
    \large
    \begin{tabular}{ccccc}
    \includegraphics[width=44mm, height= 29mm]{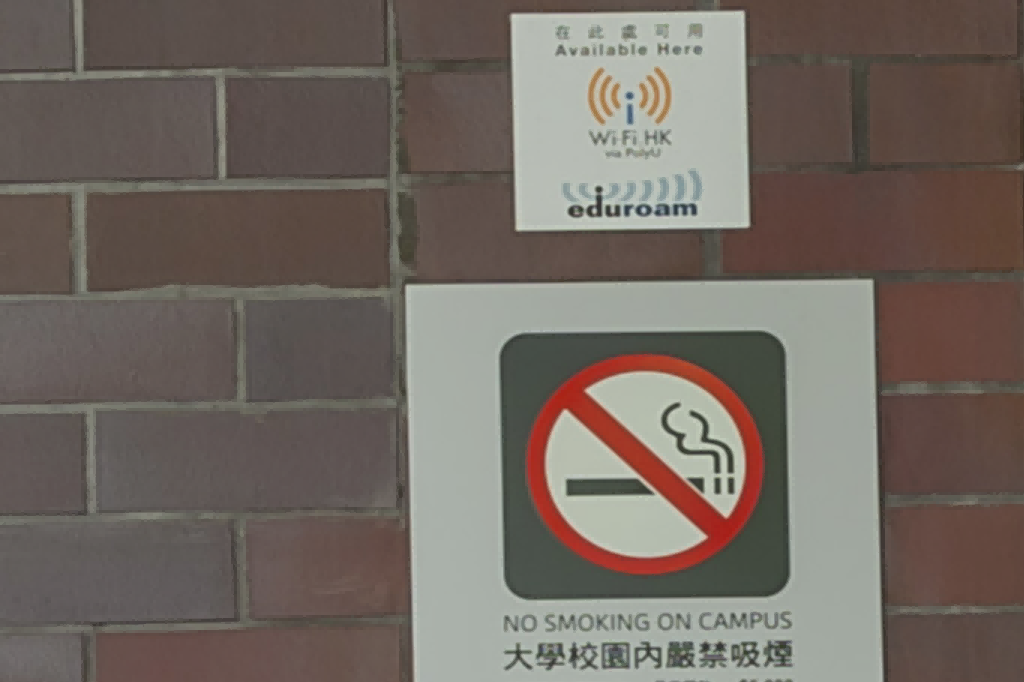}  &
    \includegraphics[width=44mm, height= 29mm]{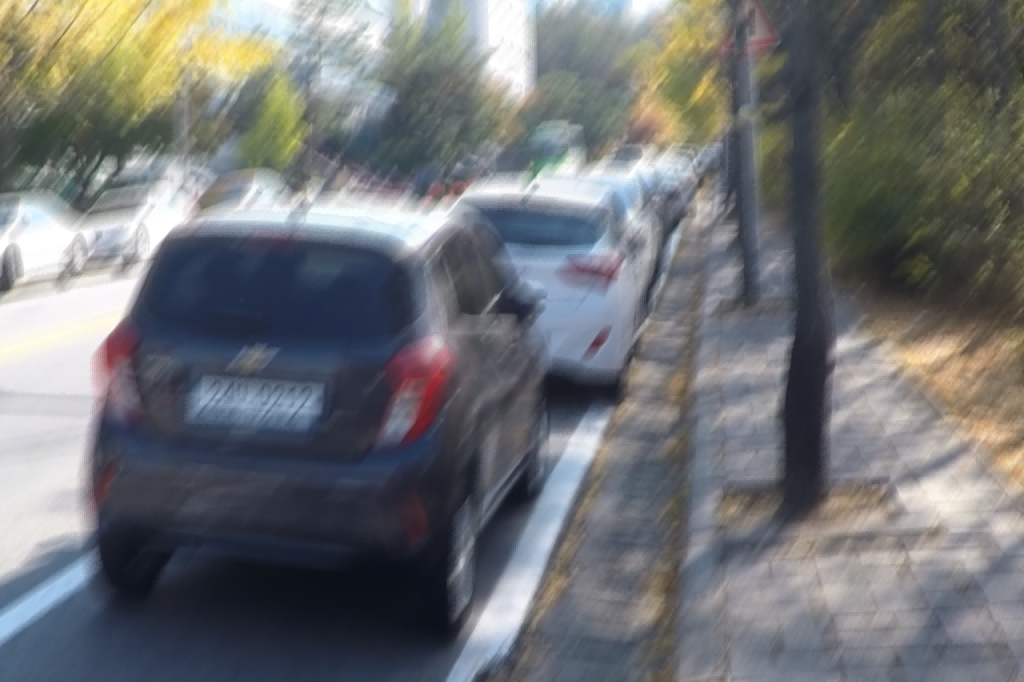}  &
    \includegraphics[width=44mm, height= 29mm]{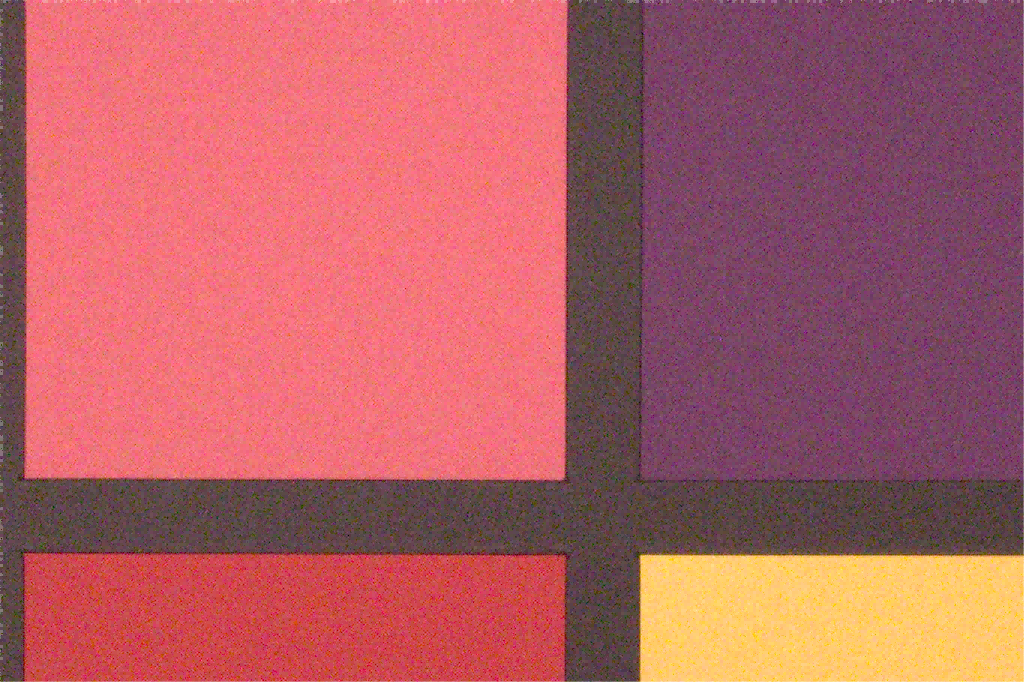}  &
    \includegraphics[width=44mm, height= 29mm]{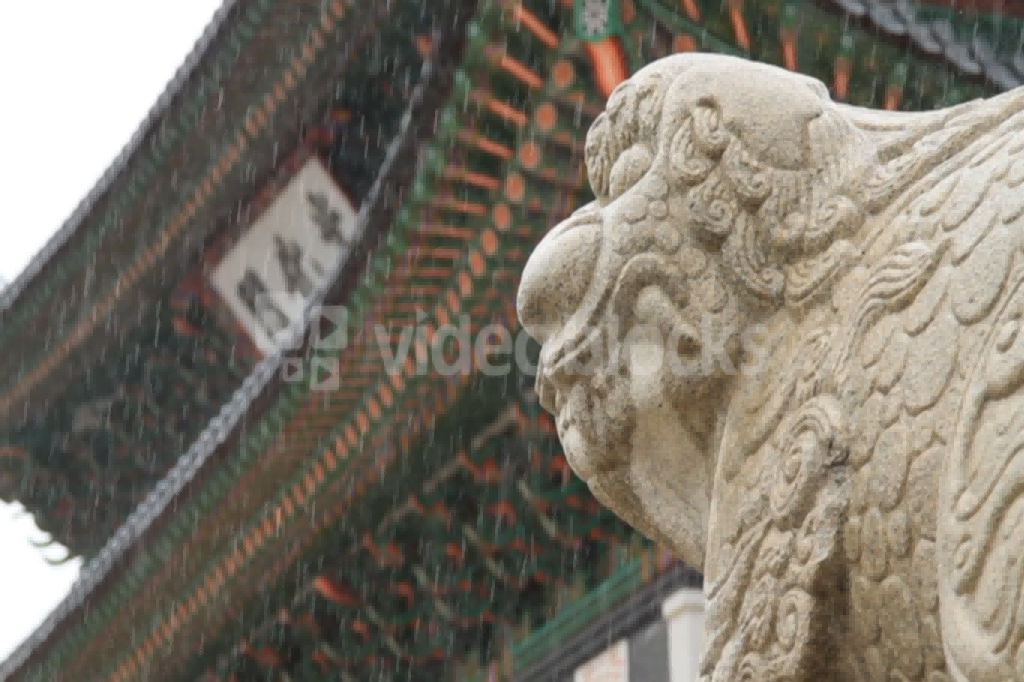}  &
    \includegraphics[width=44mm, height= 29mm]{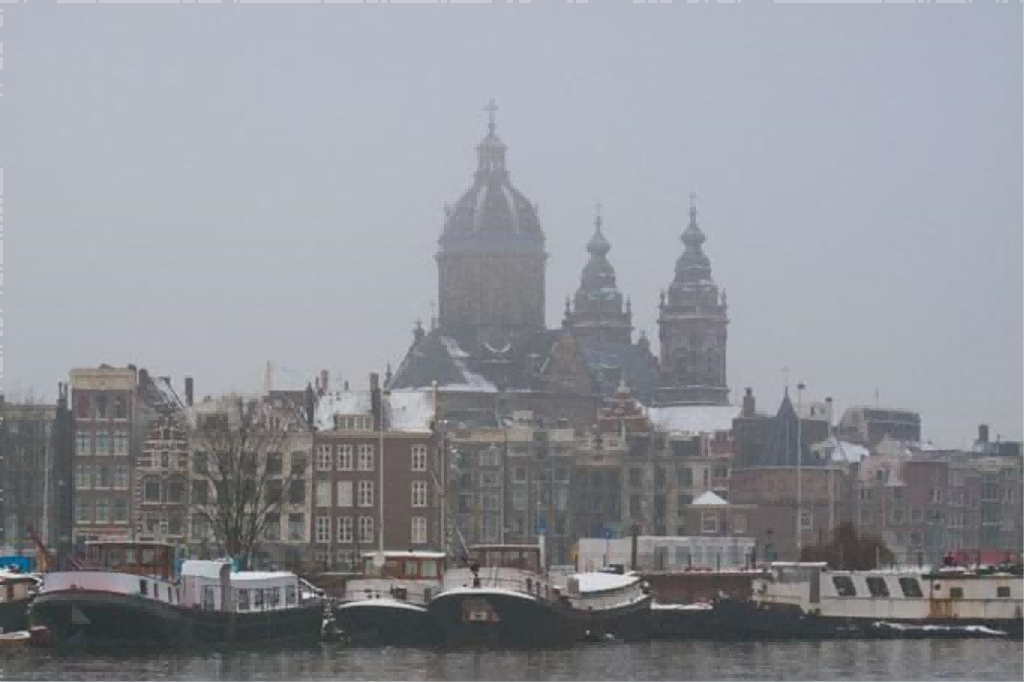} 
    \vspace{1mm}
    \\
    
    Resize  &
    Blur  &
    Noise &
    Rain  &
    Haze
    \vspace{1mm}
    
    \\
    \end{tabular}
    \end{adjustbox}
\\
\begin{adjustbox}{valign=t}
    \large
    \begin{tabular}{ccccc}
    \includegraphics[width=44mm, height= 29mm]{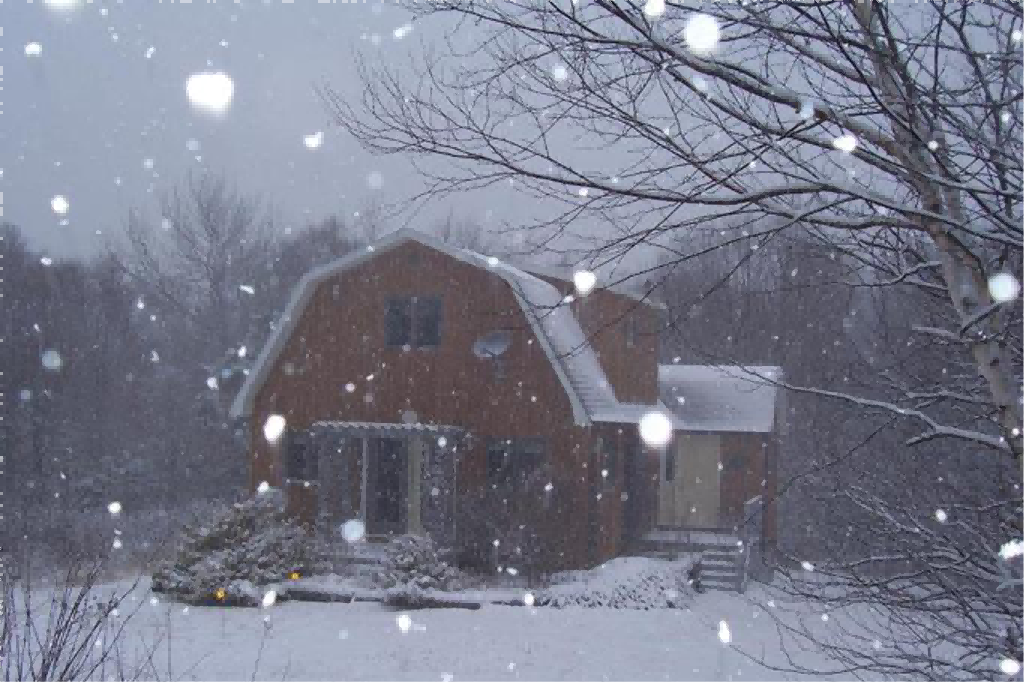}  &
    \includegraphics[width=44mm, height= 29mm]{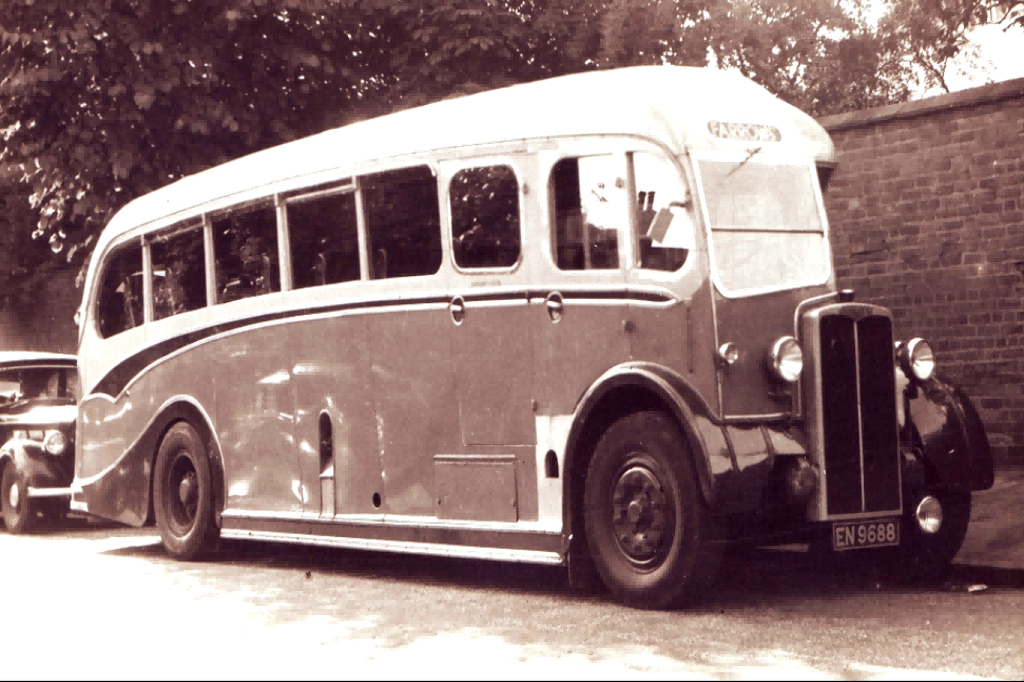}  &
    \includegraphics[width=44mm, height= 29mm]{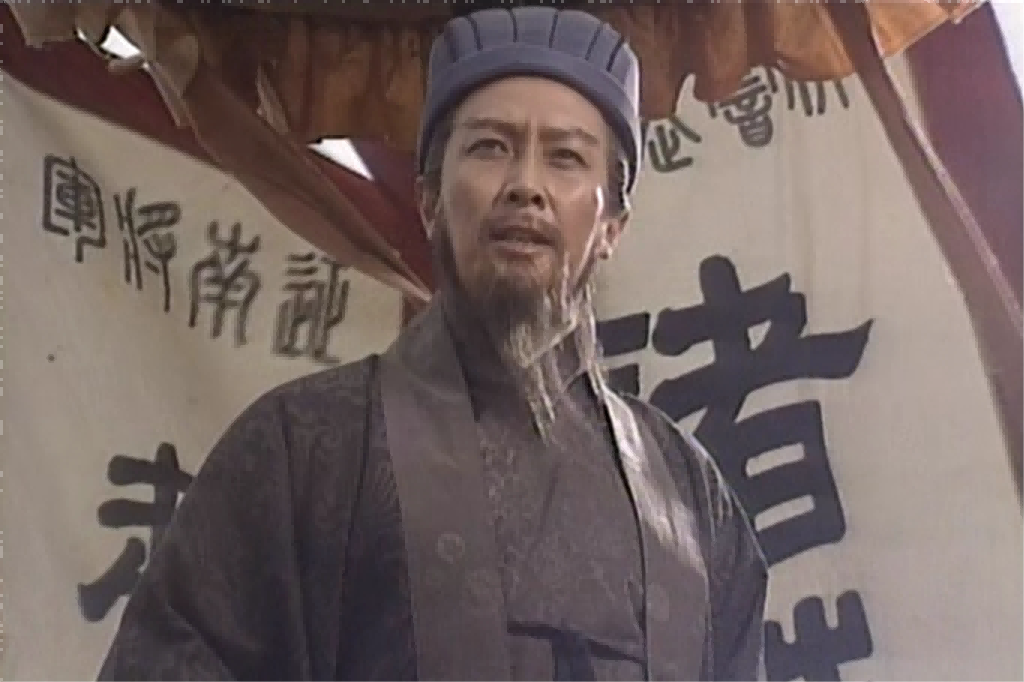}  &
    \includegraphics[width=44mm, height= 29mm]{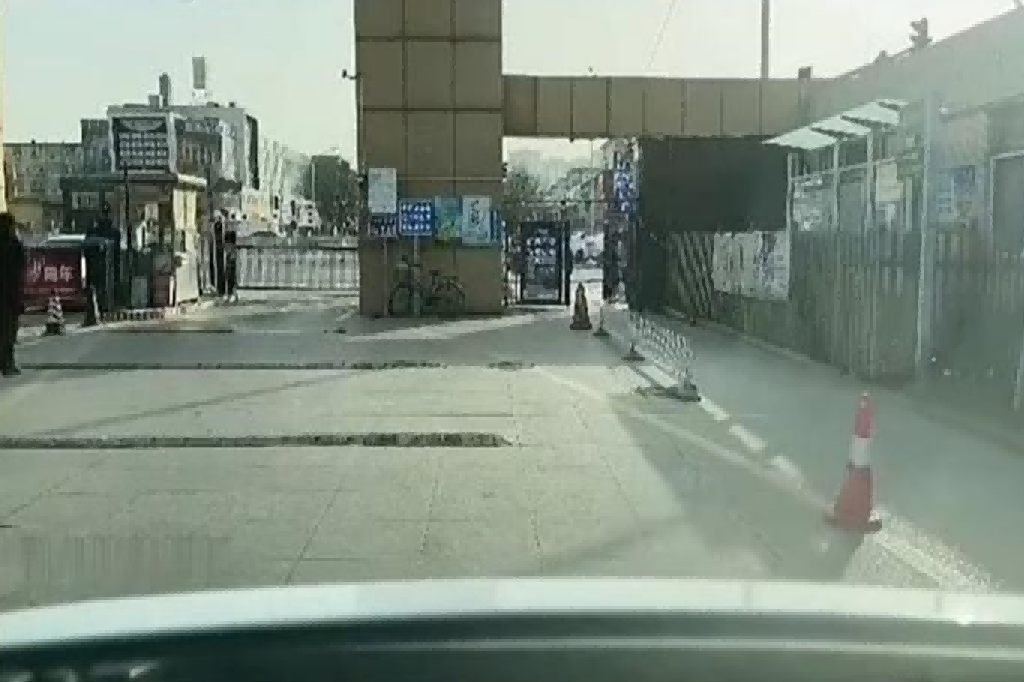}  &
    \includegraphics[width=44mm, height= 29mm]{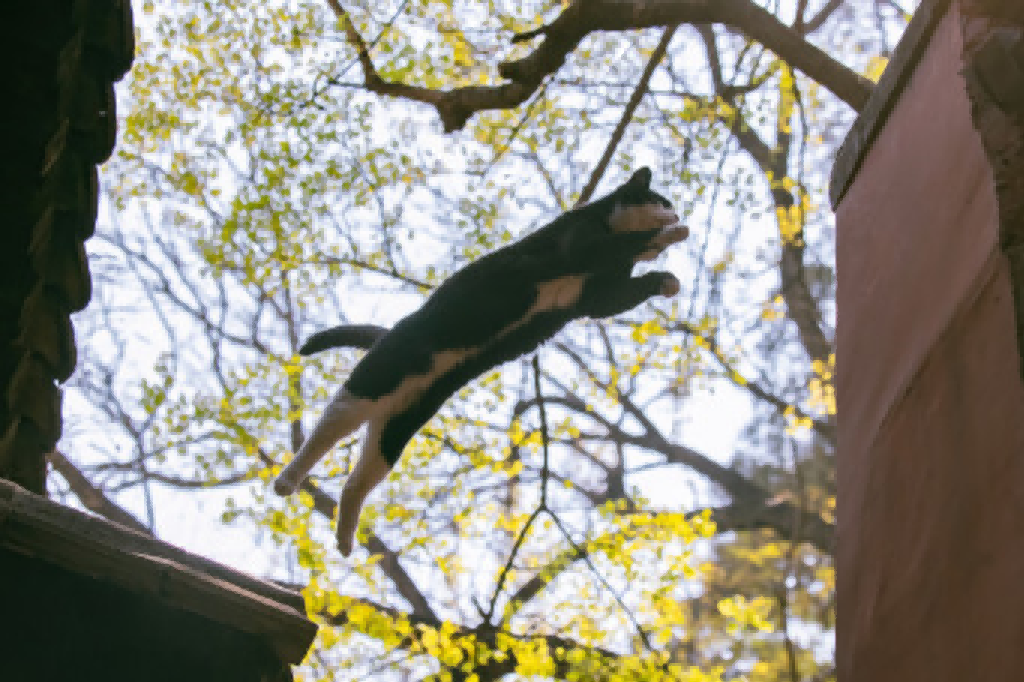} 
    \vspace{1mm}
    \\
    
    Snow  &
    Old Photos  &
    Old films &
    Surveillance  &
    Low-Quality \\ 
    & & & & Internet Image
    \vspace{1mm}

    \\
    \end{tabular}
    \end{adjustbox}
\end{tabular}}
\vskip -0.3cm
\caption{The demo of real world images. They contain 10 types with 100 real-world testing images per type.}
\label{fig:real}
\vskip -0.6cm
\end{figure*}

3) \emph{Select representative tasks for testing.}
The combination of degradations makes it impractical for us to test on all tasks.
In fact, due to resource constraints, we can only test on very limited representative tasks.
Also, we cannot arbitrarily select a small number of tasks for testing, because a few samples are unable to represent the whole degraded image space and one may only select cases in their favour to compare, resulting in nonobjective and unfair comparison.
We propose to select a subset of representative test tasks (or degradation types), which have large inter-task distances, to cover a wide range of degradation space.
We first include the above 10 single tasks with selected parameters.
Second, for mixture tasks, we sample a large number of possible mixture tasks following Eq.~\eqref{eq:degradations} with order $k$.
We conduct spectral clustering analysis~\cite{ng2001spectral,von2007tutorial} on these generated mixture tasks as different degradation combinations may lead to similar visual results.
We select the clustering centers as the representative testing tasks.
The detailed process is in the appendix.

The order $k$ in our experiments samples from $2$ to $5$.
For each $k$, ten representative tasks are collected.
As a result, we select 50 representative tasks including the single tasks.
In addition to 50 representative tasks, we collect the other 50 randomly generated degradations following Eq.~\eqref{eq:degradations} for each order from $k=1$ to $k=5$ for testing and cross-validation.
The differences in model performance between the clustering center set and the cross-validation set can often indicate generalization issues.
These 100 degradations/tasks cover a wide range of testing scenarios.
The detailed degradation formulations, parameters and visual demos are shown in the appendix.


\paragraph{Real-world test images.}
\label{Real-world}
Evaluating real-world scenarios is a feasible way to measure $\mathcal{R}$ of Eq.~\eqref{eq:GIR-opt} which represents the situations that can not be sampled.
Therefore, we also collect 10 types of real-world testing images.
For some of the single tasks involved, there has been a lot of works proposing real-world datasets for them.
GIR should show better results on these datasets.
As shown in Figure~\ref{fig:real}, we include the RealSR dataset \cite{cai2019toward} for task \texttt{resize}, the GoPro \cite{nah2017deep} and the RealBlur \cite{rim2020real} dataset for task \texttt{blur}, and the SIDD \cite{SIDD} for \texttt{noise}.
We can access the ground truth for reference in these test sets.
We also include the SPA dataset \cite{wang2019spatial} for \texttt{rain}, \cite{RESIDE} for \texttt{haze} and \cite{liu2018desnownet} for \texttt{snow}.
For these data, we have no ground truth as the reference.
We also collected additional degradation types that are not present in previous studies, \ie old photos, old films, surveillance and low-quality Internet images.
These images without corresponding ground truth mostly come from the Internet, with a small part from \cite{WuLimYang13} and \cite{wan2020bringing}. 
For the real-world testing images, we use the non-reference IQA methods such as NIQE as metrics to reflect the reconstruction performance.

\subsection{Evaluating Model Generality}
\label{sec:generality}

To evaluate the performance of MIR and BIR, previous works~\cite{liang2022efficient,kong2022reflash} directly calculate the average performance on a randomly-sampled small-sized test set.
However, we note that the average PSNR is not entirely sufficient for GIR because a higher average PSNR may be obtained by performing well on only a portion of the tasks, without maintaining model generality.
%
We want to achieve a balanced and acceptable performance across various tasks that GIR can cover, instead of focusing on only a part of tasks and sacrificing performance on the other tasks.
Thus, in Eq.~\eqref{eq:GIR1}, the generality term $\mathcal{G}$ is added to the evaluation of GIR.
Here we give a simple and reasonable choice of $\mathcal{G}$.

As different degradations may have different quantitative results (\eg, PSNR), absolute numbers are no longer reliable indicators.
The average quantitative results across all testing samples will lead to misleading comparison
results for GIR evaluation. 
We need baselines and hope that they can describe the performance expectations of using a simple deep model on a task, and the performance we can expect to obtain when using a relatively good deep model.
Similar to SEAL~\cite{zhang2023seal}, we introduce an acceptance line and an excellence line to evaluate the overall performance of GIR.
For every single task $t$, we train a three-layer convolutional deep network~\cite{SRCNN} supervised without any multi-task setting.
Then we record its testing loss $\hat{\mathcal{L}}^t(\theta^{\mathrm{AC}})$, defining this as the acceptance line.
For the excellence line, we replace the shallow network with a more advanced deep residual network~\cite{ResNet}.
The performance of this network is recorded as the excellence line $\hat{\mathcal{L}}^t(\theta^{\mathrm{EXC}})$.
The generality is measured by comparing the GIR model with the acceptance line and excellence line and see on how many tasks GIR can meet our requirements for acceptance and excellence.
These ratios are also called the acceptance ratio (AR) and the excellence ratio (ER) as
\begin{align}
    \mathrm{AR}&=\frac{1}{T^{\mathrm{Test}}}\sum_{t=1}^{T^{\mathrm{Test}}} \mathbbm{1} \{\hat{\mathcal{L}}^t(\theta^{\mathrm{GIR}})\geq\hat{\mathcal{L}}^t(\theta^{\mathrm{AC}})\},\\
    \mathrm{ER}&=\frac{1}{T^{\mathrm{Test}}}\sum_{t=1}^{T^{\mathrm{Test}}} \mathbbm{1} \{\hat{\mathcal{L}}^t(\theta^{\mathrm{GIR}})\geq\hat{\mathcal{L}}^t(\theta^{\mathrm{EXC}})\},
\end{align}
where $\mathbbm{1}\{\cdot\}$ is the indicator function.
Specifically, for the 100 select testing tasks, we train 100 models for the acceptance line and 100 models for the excellent line.
%
The generalizability and overall performance of the GIR model can be effectively measured by leveraging the relative referencing metrics of AR and ER across the degradation space.
\begin{figure*}[t]
  \centering
  \includegraphics[width=\linewidth]{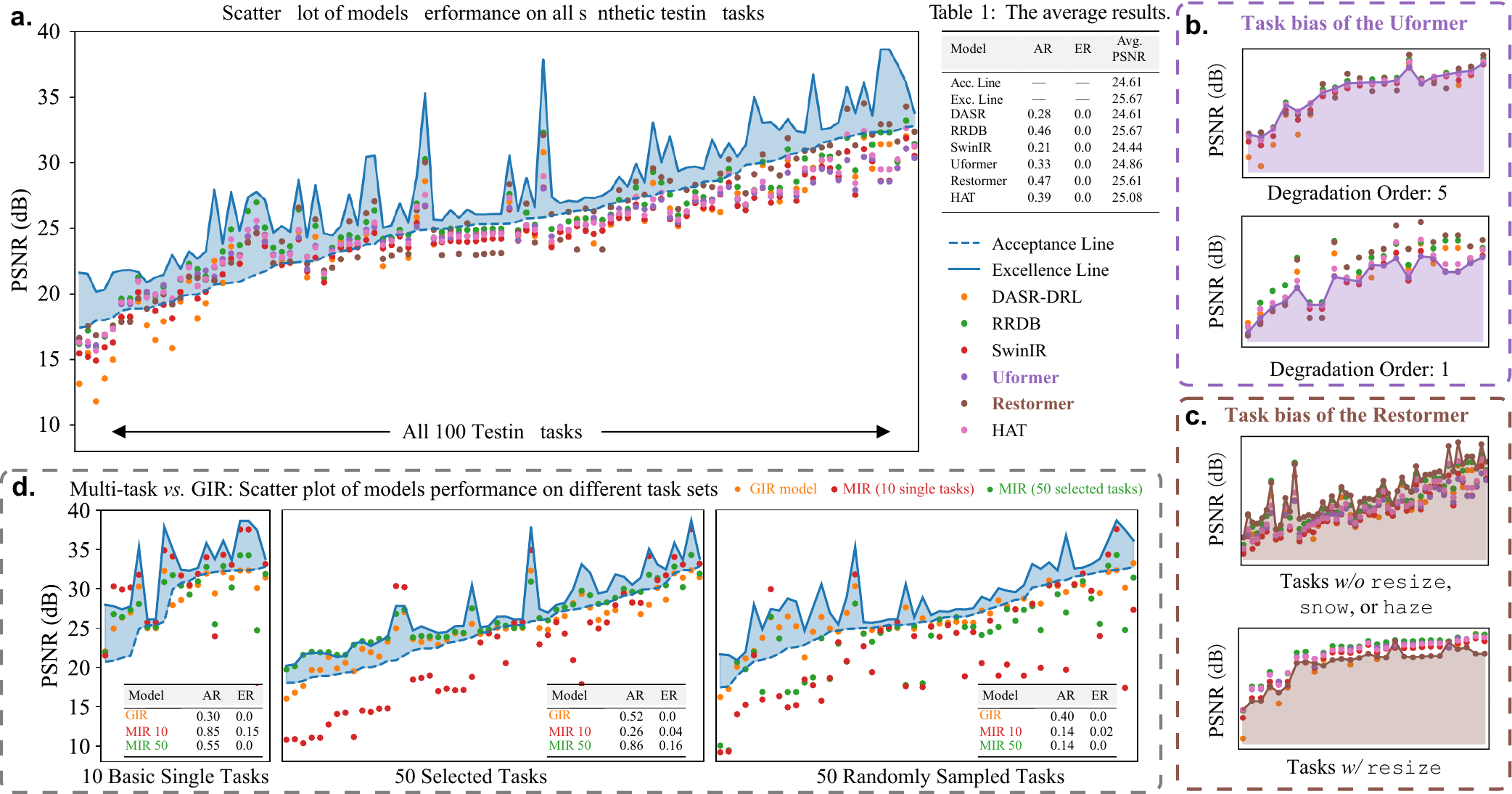}
  \caption{Benchmark results of the existing approaches using the GIR evaluation protocol. (a) shows the scatter plot of models performance on all synthetic testing tasks; (b) shows the task bias of Uformer and (c) shows the task bias of Restormer, please refer to Section~\ref{sec: Practical Difficulties} for more details; (d) shows the performance of MIR-10, MIR-50 and GIR models on different degradations respectively.}
  \label{fig:baseline}
\end{figure*}

\paragraph{Discussion.}
In this section, we propose an evaluation framework for GIR problem, including 110 test tasks (100 synthetic and 10 real tasks) and two new evaluation metrics AR and ER.
%
%
In terms of data, this approach is more representative of model performance across the entire degradation space.
Previous methods usually use the results of a few selected degradation to represent the performance \cite{kong2022reflash,BSRGAN,AirNet}.
In terms of metrics, previous methods often directly compare the PSNR values.
However, our evaluation framework incorporates generalization into consideration.
ER/AR not only allows a fair comparison between GIR models but also measures the overall progress of GIR development.
When ER/AR is close to 1, we can tell that a GIR model could be comparable to simple/common single-task models.
Of course, even both ER/AR has reached 1, it does not demonstrate that GIR is realized.
But under the current conditions, increasing ER and AR should naturally encourage the models to improve their generalization ability and move toward GIR.


\section{Benchmarking Existing Approaches}
\label{sec: Benchmarking Existing Approaches}
By formally defining the task of GIR and establishing a comprehensive evaluation protocol, we can conduct a comprehensive benchmarking of various existing methods that could potentially be adapted or leveraged for the GIR task.
We first investigate the existing approaches and select the methods that meet the model principle in Section~\ref{sec: General Image Restoration}.
Then we refine and retrain them based on the GIR dataset.
Note that, the retraining process employs a straightforward strategy across all models, without incorporating any specialized techniques or tricks.
Finally, we show the performance of these methods and analyze the experimental results.
The purpose of the benchmark is not to propose a new method, but rather: (1) to verify the feasibility and importance of GIR; (2) to demonstrate the practical difficulties of GIR at the current stage and emphasize the importance of generalization; and (3) to identify potential future research directions by exploring existing approaches.
%

\subsection{Implementation}
\label{sec: Method Investigation and Implementation}

%
\paragraph{Method investigation.} 

Due to the novelty of GIR, we do not have an off-the-shelf method that can be directly applied.
Fortunately, we have learned the basic principles of choosing a method according to the discussion in Section~\ref{sec: General Image Restoration}.
Because the degradation information is not available for a GIR model, most methods that require auxiliary information input, such as, conditional restoration models \cite{SRMD,he2019modulating,he2021interactive} are ruled out.
Similarly, methods that involve pre-training on a large dataset followed by fine-tuning on specific tasks~\cite{IPT,EDT,DegAE}, or a mixture of expert methods, also cannot meet our requirements.
Because it is impractical to assign task-specific parameters to different tasks due to a large number of tasks.
We can only assume a large model with shared parameters across tasks and aim to build a fully ``blind'' model.

As mentioned in Section~\ref{sec: General Image Restoration}, the most current blind restoration methods cannot be directly applied to GIR.
However, we find some exceptions.
DASR~\cite{DASR-DRL} does not assume an explicit degradation model and learns degradation information by contrastive learning which can theoretically cover any type of degradation.
Recent works BSRGAN~\cite{BSRGAN} and RealESRGAN~\cite{RealESRGAN} provide a simple and practical way to directly integrate a large number of tasks in the training of a powerful image-to-image network.
We find that this approach can be directly extended to GIR.
Finally, we select DASR~\cite{DASR-DRL} and several representative powerful backbones including RRDB~\cite{ESRGAN}, SwinIR~\cite{liang2021swinir}, Restormer~\cite{Restormer}, Uformer~\cite{wang2022uformer} and HAT~\cite{HAT} to be benchmarked and the training method of BSRGAN~\cite{BSRGAN} and RealESRGAN~\cite{RealESRGAN} is employed to retrain these backbones.

%
\paragraph{Training strategy.}
For the training strategy, we use a training strategy similar to BSRGAN~\cite{BSRGAN} and RealESRGAN~\cite{RealESRGAN}.
We synthesize the training image pairs according to the method described in Eq.~\eqref{eq:degradations}. 
During training, we sample the clean ground truth images from DIV2K~\cite{DIV2K} and Flickr2K~\cite{timofte2017ntire} datasets, including 3,450 2K images in total.
We generate images with \texttt{rain}, \texttt{haze} and \texttt{snow} offline because these degradations need global information about the images to adjust parameters or estimate depth.
The remaining degradations are applied to image patches on the fly with a wide range of random parameters.
During training the models, $l_1$-norm loss function~\cite{L1} is adopted with the Adam optimizer~\cite{ADAM} ($\beta_1$ = 0.9, $\beta_2$ = 0.999). 
The batch size is set to 8, and the patch size is $128\times128$. 
The cosine annealing learning strategy is applied to adjust the learning rate.
The initial learning rate is $2\times10^{-4}$ and decays to $10^{-7}$. 
The period of cosine is 500k iterations. 
For DASR, we first train the degradation encoder using the same training data and then train the main module following the original paper.
All models are built on the PyTorch~\cite{pytorch}.



\subsection{Effectiveness of GIR}
\label{sec: Superiority of GIR}
In this section, we show the effectiveness and importance of GIR through the quality and quantity results of benchmarked methods.
We only use the simplest training strategy without any generalization optimization to train the existing models. 
Nonetheless, these models demonstrate better visual quality and generalization compared to MIR and progressive single-task results.

\paragraph{Generalization ability.}
%
First, we compare the GIR models with multi-task solutions to illustrate the generalization ability of GIR models. 
This comparison is intended to evaluate the generality $\mathcal{G}$.
We train two multi-task image restoration models ``MIR-10'' and ``MIR-50'' for comparison.
The MIR-10 and MIR-50 are trained using 10 basic single / 50 representative tasks with the RRDB \cite{RealESRGAN} backbone.
\figurename~\ref{fig:baseline}{\color[HTML]{FF0000}d} shows the performance and AR/ER of MIR-10, MIR-50 and GIR models on different degradations respectively.
The GIR model also uses the same RRDB backbone.
As can be seen, the MIR-10 model obtains the highest PSNR value on the ten single tasks but performs poorly on almost all the remaining tasks, illustrating poor generalization performance and generality.
The MIR-50 model, similarly, performs well on the fifty selected tasks on which it is trained.
The performance also drops dramatically when it is tested on the other fifty randomly generated degradations.
%
%
It shows that although the MIR models can achieve acceptable results on their training tasks, they can hardly generalize to other similar tasks.
Even a simple change in the degradation strength or order can cause the MIR models to fail.
But the simple GIR model can achieve better generalization.

\begin{figure*}[t]
\scriptsize
\centering
\resizebox{1\textwidth}{!}{
\begin{tabular}{cc}
\\
\hspace{-0.4cm}

\begin{adjustbox}{valign=t}
\begin{tabular}{ccccccc}
\includegraphics[width=0.145\textwidth]{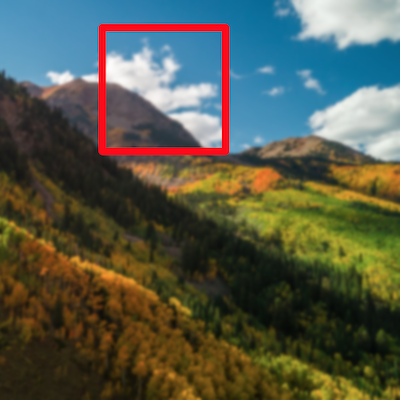} \hspace{-3.5mm} &
\includegraphics[width=0.145\textwidth]{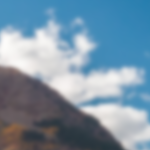} \hspace{-3.5mm} &
\includegraphics[width=0.145\textwidth]{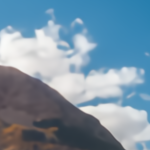} \hspace{-3.5mm} &
\includegraphics[width=0.145\textwidth]{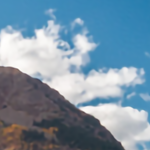} \hspace{-3.5mm} &
\includegraphics[width=0.145\textwidth]{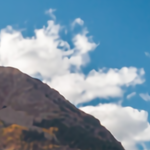} \hspace{-3.5mm} &
\includegraphics[width=0.145\textwidth]{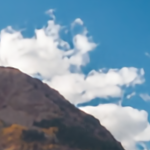} \hspace{-3.5mm} &
\includegraphics[width=0.145\textwidth]{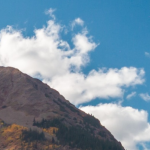} \hspace{-3.5mm} 
\\
\end{tabular}
\end{adjustbox}
\\
\hspace{-0.4cm}
\begin{adjustbox}{valign=t}
\begin{tabular}{ccccccc}
\includegraphics[width=0.145\textwidth]{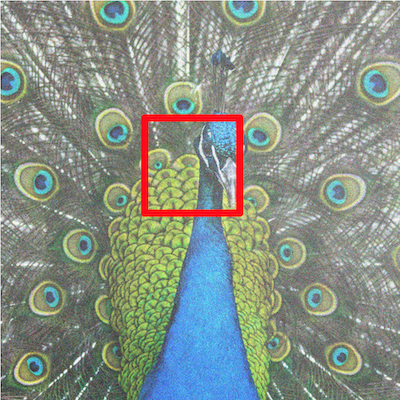} \hspace{-3.5mm} &
\includegraphics[width=0.145\textwidth]{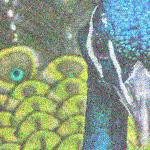} \hspace{-3.5mm} &
\includegraphics[width=0.145\textwidth]{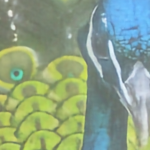} \hspace{-3.5mm} &
\includegraphics[width=0.145\textwidth]{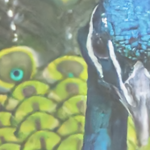} \hspace{-3.5mm} &
\includegraphics[width=0.145\textwidth]{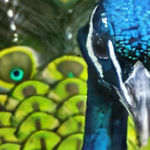} \hspace{-3.5mm} &
\includegraphics[width=0.145\textwidth]{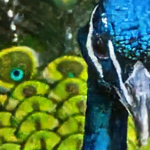} \hspace{-3.5mm} &
\includegraphics[width=0.145\textwidth]{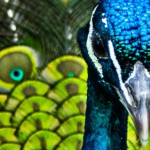} \hspace{-3.5mm} 
\\
\end{tabular}
\end{adjustbox}
\\
\hspace{-0.4cm}

\begin{adjustbox}{valign=t}
\begin{tabular}{ccccccc}
\includegraphics[width=0.145\textwidth]{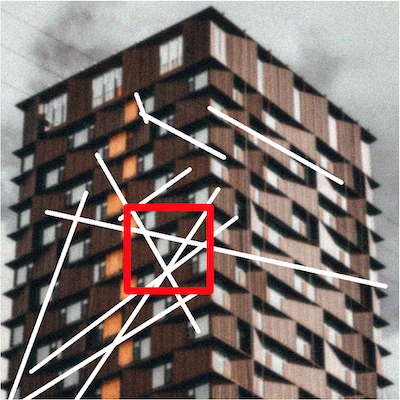} \hspace{-3.5mm} &
\includegraphics[width=0.145\textwidth]{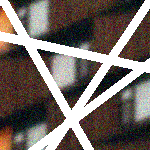} \hspace{-3.5mm} &
\includegraphics[width=0.145\textwidth]{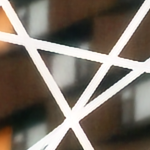} \hspace{-3.5mm} &
\includegraphics[width=0.145\textwidth]{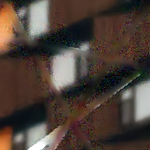} \hspace{-3.5mm} &
\includegraphics[width=0.145\textwidth]{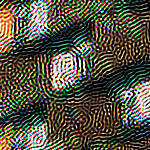} \hspace{-3.5mm} &
\includegraphics[width=0.145\textwidth]{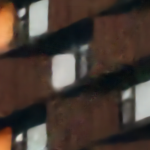} \hspace{-3.5mm} &
\includegraphics[width=0.145\textwidth]{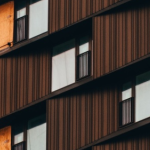} \hspace{-3.5mm} 
\\
\hspace{-4mm} &
Degraded image \hspace{-4mm} &
RealESRNet \cite{RealESRGAN} \hspace{-4mm} &
MIR-10 \hspace{-4mm} &
Step-by-step \hspace{-4mm} &
GIR (RRDB) \hspace{-4mm} &
Ground Truth
\\
\end{tabular}
\end{adjustbox}
\\
\end{tabular}}
\caption{Visual comparison on the presented synthetic testing set. Zoom in for a better view.}
\label{Fig:CAR}
\end{figure*}

\paragraph{Visual Effects.}
Next, we present visual comparisons between the GIR model and single-task as well as mixture-task models.
For the compared models, we first include the RealESRNet \cite{RealESRGAN} as it is known to deal with multiple degradations.
Then we use a multi-task image restoration model ``MIR-10'' for testing.
As we include the mixture degradations, we also employ an intuitive solution for testing.
``Step-by-step'' represents processing images using non-blind models according to certain degradations (\eg, if an image is degraded first by haze and then by blur, the ``Step-by-step'' solution will first deblur and then dehaze).
Without joint training between different non-blind models, this shows how the discrete single-task models handles mixture degradations.
We also test the GIR model with the RRDB backbone.
\figurename~\ref{Fig:CAR} shows the visual comparison of different methods.
The first row is a case with a single task \texttt{blur}. Almost every model generates good results except that RealESRNet generates over-smooth textures.
Individual tasks are generally within the scope of the existing methods.
The second row is a mixture case degraded first by haze and then by noise, which can happen in real life.
Both RealESRNet and MIR-10 fail in this case.
Although MIR-10 was trained with both \texttt{haze} and \texttt{noise}, it only removes noise and cannot deal with this mixture task.
The ``Step-by-step'' can handle such a simple mixture.
The third row shows a more complex mixture task, containing \texttt{damage}, \texttt{noise}, \texttt{ringing}, and \texttt{blur}.
Due to the accumulated errors and artifacts \cite{Path-restore}, even if we know the specific degradations, ``Step-by-step'' still can not produce stable results.
Other existing methods unsurprisingly fail.
For all the cases, the simple GIR model can output reasonable results.

\begin{figure}[t]
\centering
\includegraphics[width=0.98\linewidth]{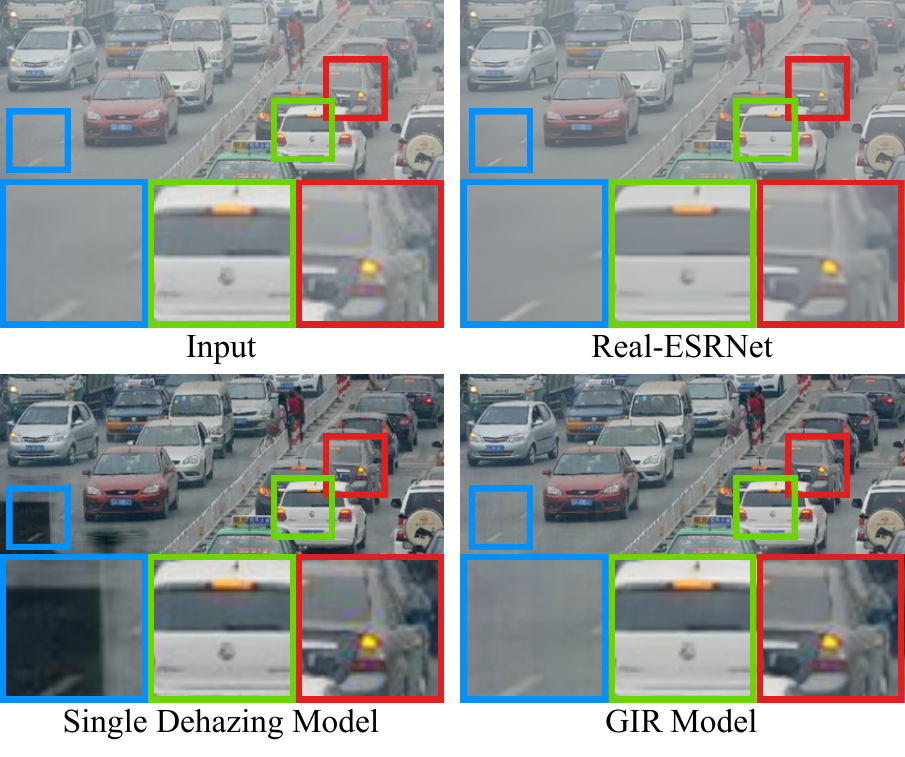}
\caption{Comparison on the real-world testing image. Zoom in for a better view.}
\label{fig:real}
\end{figure}

\paragraph{Real-world cases.}
Finally, we show the reconstruction results of a real-world image in \figurename~\ref{fig:real}.
It focuses on the evaluation of $\mathcal{R}$.
The input image contains observable haze, noise and compression artifacts.
The RealESRNet model removes almost all the artifacts but can do nothing for the haze.
Compared to the GIR model, the dehaze model has similar dehazing performance but also retains unnatural artifacts and noise.
The GIR model not only removes the haze, but also fixes various artifacts in the image.
Quantitative results also speak to the same conclusion.
The degradations involved in these four test sets in Table~\ref{table:NIQE} have not been extensively studied, so few works can address them simultaneously.
We show the results of the deblur and denoise models, as these two degradations are the most common ones.
Besides that, we also compare our GIR models with RealESRNet.
The results illustrate the GIR models can obtain better NIQE values in almost all these datasets.

\begin{table}[]
\centering
\resizebox{\linewidth}{!}{
\begin{tabular}{lcccc}
\toprule
\rowcolor{color3} Real Data                           & Deblur                     & Denoise & RealESRNet & GIR          \\
\midrule
\texttt{Resize}      & {\color[HTML]{FF0000} 5.30} & 6.29     & 6.03       & 5.65                        \\
\texttt{Blur}       & 5.96                        & 6.28     & 6.71       & {\color[HTML]{FF0000} 5.68} \\
\texttt{Noise}      & 6.37                        & 6.14     & 9.18       & {\color[HTML]{FF0000} 5.30} \\
\texttt{Rain}       & 6.53                        & 6.74     & 6.45       & {\color[HTML]{FF0000} 5.85} \\
\texttt{Haze}      & 5.71                        & 5.59     & 6.57       & {\color[HTML]{FF0000} 4.87} \\
\texttt{Snow}       & 5.00                        & 4.79     & 5.10       & {\color[HTML]{FF0000} 4.67}  \\
\texttt{Old film}       & 6.71                        & 6.90     & 6.72       & {\color[HTML]{FF0000} 6.45}  \\
\texttt{Old photo}      & {\color[HTML]{FF0000} 3.48}                       & 3.69     & 4.71       &  3.62  \\
\texttt{Surveillance}       & 6.28                        & 6.58     & 6.67       & {\color[HTML]{FF0000} 6.03}  \\
\texttt{Internet}       & 5.28                        & 5.28     & 4.85       & {\color[HTML]{FF0000} 4.84}  \\
\bottomrule
\end{tabular}}
\caption{The NIQE performance of different models on real test sets.}
\label{table:NIQE}
\end{table}

\paragraph{Significance of GIR.}
\begin{table*}[t]
\centering


\subfloat[Ablation study of training data scales.\label{abl:1}]{
\scalebox{0.76}{
\begin{tabular}{ll|c|c|c}
    \toprule
    \rowcolor{color3} ID & Data Scale   & AR & ER & Avg. PSNR \\ \midrule
    (a) & Quarter of DF2K  & 0.49 & 0.0 &  25.69   \\ 
    (b) & Half of DF2K & 0.42 & 0.0 &   25.51     \\ 
    (c) & All of DF2K   & 0.46 & 0.0 &   25.67  \\ 
    (d) & ImageNet + DF2K & 0.44 & 0.0 & 25.38   \\ \bottomrule
  \end{tabular}
}}
\subfloat[Ablation study of training patch size.\label{abl:2}]{
\scalebox{0.76}{
\begin{tabular}{ll|c|c|c}
    \toprule
    \rowcolor{color3} ID & Patch size & AR & ER & Avg. PSNR \\ \midrule
    (e) & 64  &  0.29 & 0.0 &  24.84  \\ 
    (f) & 128 &  0.49 & 0.0 &   25.67     \\ 
    (g) & 192 &  0.52 & 0.0 &   25.95    \\
    (h) & 256 &  0.56 & 0.03 & 26.21  \\ \bottomrule
  \end{tabular}
}}
\subfloat[Ablation study of training batch size.\label{abl:3}]{
\scalebox{0.76}{
\begin{tabular}{ll|c|c|c}
    \toprule
    \rowcolor{color3} ID & Batch size & AR & ER & Avg. PSNR \\ \midrule
    (i) & 4  &  0.35 & 0.0 &  25.16  \\ 
    (j) & 8 &  0.49 & 0.0 &   25.67     \\ 
    (k) & 16 &  0.54 & 0.02 &   26.00    \\
    (l) & 32 &  0.57 & 0.03 & 26.39  \\ \bottomrule
  \end{tabular}
}}

\label{table:ablation}
\caption{Ablation studies of the factors that might influence the results.}
\end{table*}

GIR addresses a key limitation observed in previous image restoration studies: generalization.
While often overlooked, this aspect is of paramount importance.
When juxtaposed with other image restoration methods such as single-task, MIR, and BIR, GIR's distinctiveness emerges from its encompassing research paradigm.
Instead of concentrating on a handful of scenarios, GIR addresses a comprehensive range of situations.
Thus, even with simpler models, GIR exhibits applicability across a broader spectrum.
The introduction of GIR heralded a fresh perspective in image restoration research.
Beyond its academic importance, the advancement of GIR equips users with notable image processing capabilities.
This is particularly beneficial for individuals or smaller enterprises that might lack resources for bespoke solutions.
GIR's exploration has industrially invigorated the field in a holistic manner.
The aim of this research is not to perfect a single scenario or device but to offer a universally acceptable resolution.

\subsection{Practical Difficulties of GIR}
\label{sec: Practical Difficulties}

Even with all these advantages, GIR still faces many difficulties at the current stage.
Although above existing approaches are better than other methods, they are far from the GIR goal.
We explore the characteristics of these difficulties in this section.


\paragraph{Model Generality.}
Although GIR models can solve the mixture task problem to a certain extent, their generality is still involved in our concern.
These GIR models' performance evaluated by the proposed generality protocol (acceptance line and excellence line) are summarized in \figurename~\ref{fig:baseline}{\color[HTML]{FF0000}a}.
The AR, ER, and average PSNR values are shown next to the figure.
Overall speaking, these models are far from the ideal GIR model.
Most models cannot maintain acceptable performance on most tasks.
The model with the best AR is GIR-Restormer.
However, it cannot beat a 3-layer-CNN model in more than half of these degradations.
Note that the average PSNR values of GIR-RRDB, GIR-Restormer, GIR-Uformer and GIR-HAT have exceeded it of the acceptable performances.
This conflict suggests that the new metrics are significant.
AR and ER can reflect the model's generality that the average PSNR cannot describe.
Moreover, no existing model can reach the line of excellence in all testing tasks.

Moreover, in this training setting, we found that the CNN-based model RRDB outperforms some Transformer-based models and comes close to the best model, GIR-Restormer (even surpassing it in average PSNR), despite Transformers generally demonstrating superior performance over CNNs in various single tasks.
There are many possible reasons, such as Transformer are poor at GIR or the optimization is inappropriate. 
We use same learning rate and training iterations for all models but it may be not suitable for each model and degradation. 
It may also be due to some Transformer models having a preference for specific degradations.

\paragraph{Network Bias.}
One possible reason for the poor generalization (especially for Transformer methods) may be that different networks have their preferred tasks, although they have the same training settings.
This phenomenon has been found in RealSR tasks~\cite{zhang2024real}.
We illustrate this using Uformer \cite{wang2022uformer} and Restormer \cite{Restormer}.
As illustrated in \figurename~\ref{fig:baseline}{\color[HTML]{FF0000}b}, the GIR-Uformer (purple) model excels in handling complex degradations but struggles with simpler degradation scenarios. On the other hand, \figurename~\ref{fig:baseline}{\color[HTML]{FF0000}c} reveals that the GIR-Restormer (brown) exhibits poor performance on almost all degradations involving resize, yet demonstrates impressive capabilities in addressing the remaining tasks. These two Transformer-based models exhibit distinct task-specific preferences that are not observed in SwinIR and HAT. This indicates that future research on the backbone of GIR models should carefully consider task preference to achieve optimal generalization across various tasks.
We take a closer look at them in Section~\ref{sec: Understanding GIR}



\paragraph{Effect of Training Data Scale.}
General large models are often inseparable from extremely large datasets.
However, we question whether this perception holds true for GIR.
We investigate the influence of training data scale for GIR models.
Table~\ref{abl:1} shows that using half or even a quarter of all the training data can still achieve comparable performance.
It proves that the core bottleneck of current GIR is not dataset scale.
Existing models may not fully utilize even a quarter of the training data with these complex degradations, let alone larger datasets.

A recent study may offer some explanation for this anomaly.
Gu et al. \cite{gu2023networks} found that using fewer images for training may actually lead to better generalization performance in deraining studies.
This is because the network tends to choose the simpler task between learning degradation and image content.
In most cases, learning degradation is simpler than learning image content.
Our results may suggest that a better GIR model should benefit from more training data.
It also shows some of the inadequacies of today's technology.

Surprisingly, adding ImageNet~\cite{ImageNet} images to the training set decreases the GIR performance.
This confirms our conclusion that GIR needs high-quality images for training and testing, while the image quality of ImageNet is not desirable, containing noise and compression.
Therefore, high-quality training data is still in demand for GIR.

\paragraph{Effect of Training Strategy.}
We also explore the influence of the training batch and patch size.
As shown in Table~\ref{abl:2} and Table~\ref{abl:3}, we can see that all three metrics are improving with the increasing batch and patch size.
However, the marginal effect of this improvement is diminishing.
We also find that models with larger batch and patch sizes may achieve the excellence line in some degradations.
All these degradations included \texttt{snow} or \texttt{rain}, illustrating that batch and patch sizes may have a greater impact on such degradations.
Detailed results are provided in appendix.

\section{Interpret GIR Model for Future Direction}
\label{sec: Understanding GIR}
The GIR problem emphasizes the ability to generalize. The various models we tested also exhibited different generalization performances.
However, apart from the difference in PSNR of the output images, we also want to know what contributes to or hurts the network's generalization ability.
This will inspire future research.
Next, we investigate the GIR models through the lens of network interpretation and visualization.
In the following experiments, we still focus on the ``MIR-10'', ``MIR-50'' and the GIR model mentioned at the beginning of Section~\ref{sec: Superiority of GIR}.
We focus on three tasks as our research objectives:
The first task is one of the ten basic single tasks and is included in the training set of the MIR-10 and MIR-50 models.
The second task is sampled from the fifty representative testing tasks and is in the training set of the MIR-50 model.
The last task is randomly generated and not presented in any model's training set.

\paragraph{Deep Degradation Representation.}

\begin{figure}[!tbp]
    \centering
    \includegraphics[width=0.45\textwidth]{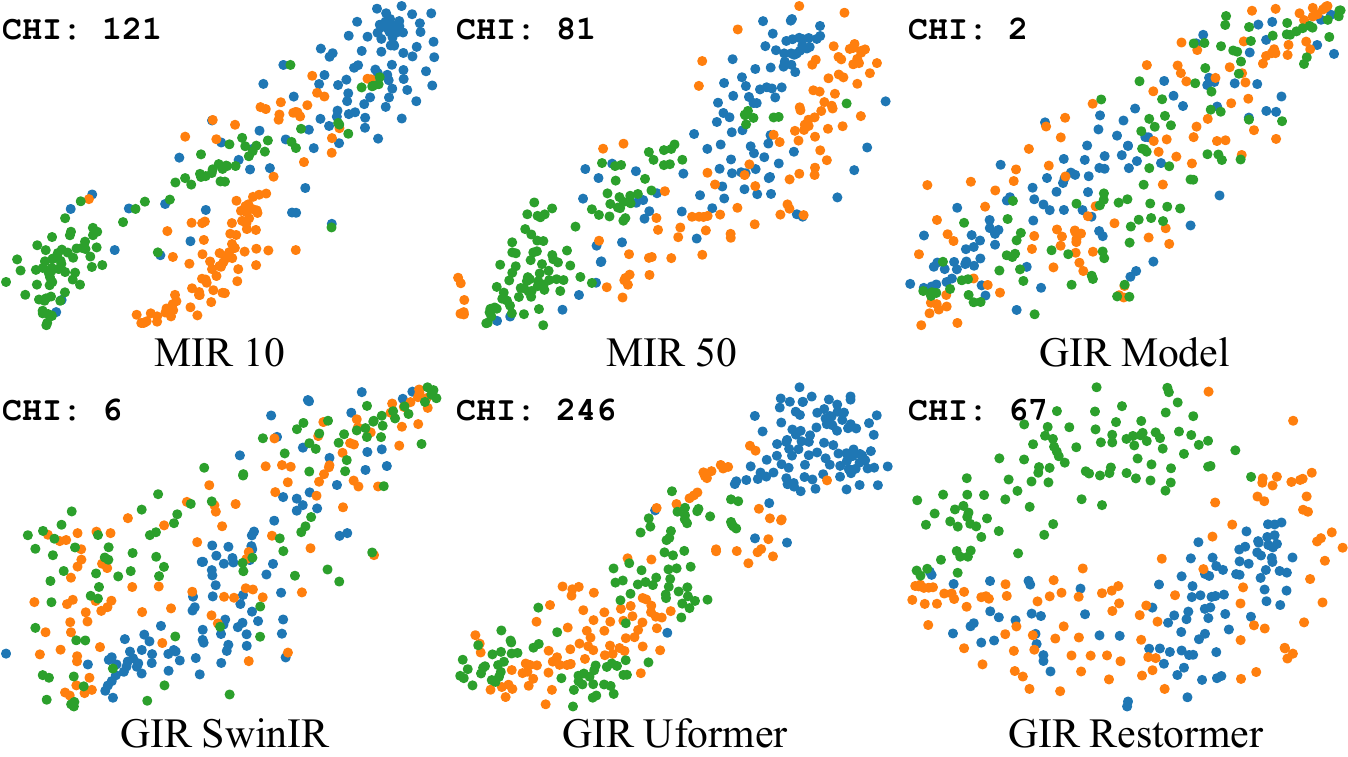}
    \caption{The DDR clustering analysis of different models.}
    \label{fig:2}
\end{figure}

Liu \etal \cite{liu2021discovering} present a concept called deep degradation representation (DDR).
Here, we will refer to Figure~\ref{fig:2} when introducing DDR.
Each point in a subfigure represents an extracted feature from an input image.
There are 300 points for every sub-figure.
These samples are produced from the above three degradations.
Each degradation corresponds to the same 100 images and is marked with the same color.
DDR reveals that networks could automatically classify the inputs to different ``degradation semantics''.
Inputs with similar degradations (points with the same color) will be clustered.
If the obtained clusters are well divided, the network tends to assign different processing strategies for different degradations, resulting in poor generalization performance.
Conversely, the network prefers to process degradations similarly and shows stable generality performance.
We can use Calinski-Harabaz Index (CHI)~\cite{calinski1974dendrite} to measure the separation degree of clusters. A lower CHI means a weaker clustering degree and indicates a better generalization ability.
From the first row of \figurename~\ref{fig:2}, we can observe the relationship between the clustering degrees from the models is MIR-10$>$MIR-50$>$GIR model.
It also illustrates that GIR models have a better generalization ability.

Besides that, the second row of \figurename~\ref{fig:2} shows that the Uformer and Restormer tend to divide input samples into clusters, while SwinIR does not show similar behaviour.
In this way, Uformer and Restormer may approach GIR by remembering training degradation combinations.
This phenomenon is related to their task preferences (See Section~\ref{sec: Practical Difficulties} -- they all try to separate individual tasks and optimize them accordingly.
This suggests that different networks have different model behaviors and produce different generality effects, even though they are all GIR models.
This deserves further study and can help us design better GIR backbone networks.

\paragraph{Attribution analysis.}

\begin{figure}[!t]
    \centering
    \includegraphics[width=0.46\textwidth]{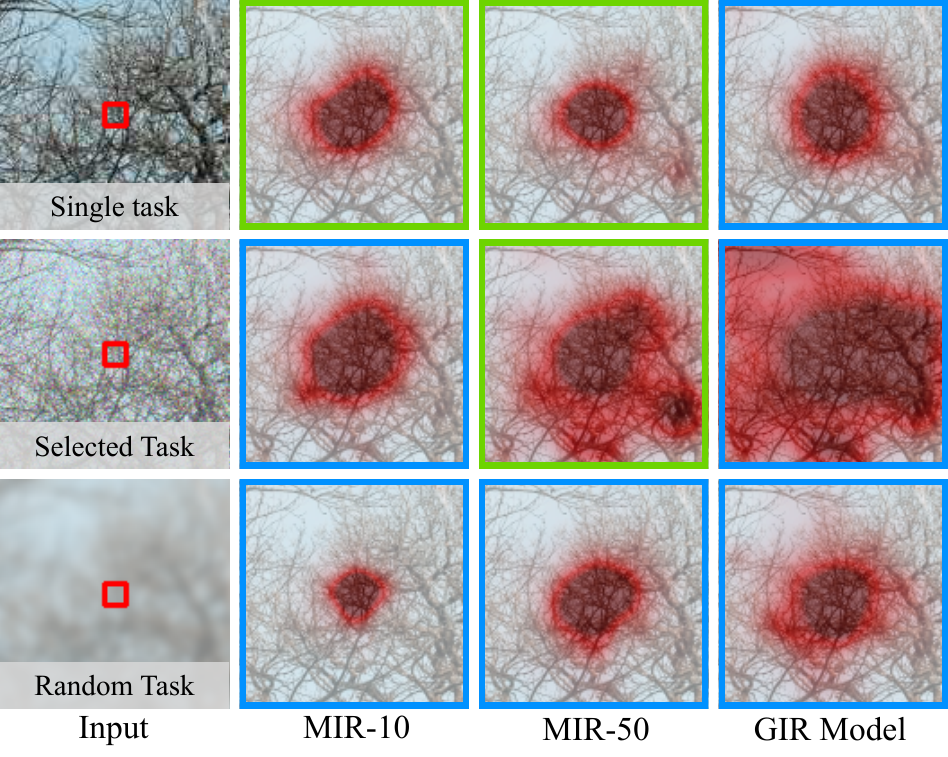}
    \caption{The attribution analysis of different models. Green boxes indicate that the task is explicitly included during model training.}
    \label{fig:3}
\end{figure}

Local attribution map (LAM)~\cite{gu2021interpreting} is a recent technique that visualizes how the network utilizes information from input pixels.
In each row of \figurename~\ref{fig:3}, three LAM maps correspond to MIR-10, MIR-50 and the GIR model, respectively.
The heat maps indicate pixels that models utilize to reconstruct the patch marked by a red box.
DI values are shown in \tablename~\ref{abl:di}. A higher DI represents a wider range of used pixels.
We can see the GIR model utilizes the most information when dealing with these tasks.
When these models are applied to the single task, all the models can obtain acceptable performance. 
These models share similar information utilization strategies and DI values.
However, the MIR models are increasingly unable to handle mixture degradations.
The pixel range used also becomes different between MIR and GIR.
MIR can only utilize more information and achieve results similar to those of GIR when faced with tasks that appeared during training (green boxes).
Compared to MIR, GIR models can directly utilize more information from the input, especially in unseen complex mixture tasks.

\begin{table}[h]
  \small %
  \begin{center}
  \scalebox{0.76}{
\begin{tabular}{lcccccc}
    \toprule
    \rowcolor{color3} Degradation   & \multicolumn{2}{c}{MIR-10} & \multicolumn{2}{c}{MIR-50} & \multicolumn{2}{c}{GIR Model} \\ 
    \rowcolor{color3} Cases & DI & PSNR & DI & PSNR &  DI & PSNR \\
    \midrule
    Single Task   & \cellcolor{green!15} 8.7 & \cellcolor{green!15} 32.37 & \cellcolor{green!15} 9.2 & \cellcolor{green!15} 31.37 & \cellcolor{blue!15} 11.3 & \cellcolor{blue!15} 30.79 \\ 
    Mixture Selected Task  & \cellcolor{blue!15} 17.8 & \cellcolor{blue!15} 12.86 & \cellcolor{green!15} 20.5 & \cellcolor{green!15} 24.8 & \cellcolor{blue!15} 22.4 & \cellcolor{blue!15} 22.73 \\ 
    Randomly Generated Task & \cellcolor{blue!15} 7.9 & \cellcolor{blue!15} 10.53 & \cellcolor{blue!15} 14.4 & \cellcolor{blue!15} 10.90 & \cellcolor{blue!15} 18.2 & \cellcolor{blue!15} 19.99 \\ \bottomrule
  \end{tabular}
}
\end{center}
\caption{The Diffusion Index of different test cases. Green boxes indicate that the task is explicitly included during model training.}
\label{abl:di}
  \end{table}


\section{Conclusion}
\label{sec: Discussion and Conclusion}

In this work, we present a new problem called general image restoration.
We have established a research framework for GIR, including its definition, dataset, and evaluation metrics, and conducted a preliminary exploration.
Our discussions demonstrate that GIR is of practical and theoretical importance.
We hope our work can lay the foundation towards a real general model in image restoration. 

\paragraph{Authors' Note.}

Initially drafted in 2022, this work represents a major step toward achieving low-level vision intelligence and generalization.
Since 2022, the development of language~\cite{GPT,GPT2,GPT3,touvron2023llama,touvron2023llama2} and multimodal models~\cite{CLIP,LLaVA,LLaVA15} has introduced new paradigms for general intelligence.
The widespread application of diffusion models~\cite{ddpm,SD} has brought new vitality to low-level vision advancements\cite{stablesr,lin2023diffbir,pasd,yu2024scaling,wu2024seesr}. 
However, even two years later, the insights provided by this paper on general image restoration remain highly valuable.

We believe that general image restoration revolves around three critical aspects: generalization, unification, and intelligence. 
The primary focus of this paper is on the issue of generalization. 
We have developed a systematic evaluation method to quantify the generalization capability of general image restoration models, which we consider fundamental to the problem. 
Unification emphasizes that general image restoration should be achieved through a unified system rather than breaking the problem into smaller, independent tasks. 
Intelligence represents the deeper requirements of general intelligence for model performance.
For example, a possible form of intelligent image restoration is that models have a better understanding of natural images space, thus they can remove degradation by removing unreasonable parts of the natural image space.

Over the past two years, large language models~\cite{GPT,touvron2023llama}, multimodal language models~\cite{CLIP,LLaVA}, and diffusion models~\cite{ddpm,SD} have each interpreted these three aspects in their respective domains. 
Multimodal language models introduce explicit intelligence driven by language, while diffusion models, based on generative priors, offer inspiration for addressing generalization issues.
The pursuit of large-scale models~\cite{gpt4,shao2021intern,cai2024internlm2} has made it possible to develop a comprehensive, universal model. 
SUPIR~\cite{yu2024scaling} is a product of this developmental path.
SUPIR erases the degradation and parts of details to produce a ``blurry input'', and then executes conditional generation based on it.
Although SUPIR utilizes powerful priors to greatly improve the visual effect of restoration, it does not solve the generalization problem of degradation outside the training set.

General image restoration points toward a research direction focused on developing unified, generalizable, and intelligent image restoration models. 
However, this direction still faces challenges related to generalization and intelligence, which will be central issues in future image restoration research.


{\small
\bibliographystyle{ieee_fullname}
\bibliography{egbib}
}

\clearpage

\renewcommand\thesection{\Alph{section}}
	\renewcommand\thesubsection{\thesection.\arabic{subsection}}
	\renewcommand\thefigure{\Alph{section}.\arabic{figure}}
	\renewcommand\thetable{\Alph{section}.\arabic{table}} 
	
	\noindent{\LARGE{\textbf{Appendix}}}
	
	\setcounter{section}{0}
	\setcounter{figure}{0}
	\setcounter{table}{0}

\vspace{0.6cm}

In this appendix, we show the details of testing and training data including basic single tasks, mixed tasks selecting methods and the combination parameters \& demo of the selected tasks. 
After that, we display more detailed results of the synthesized test set.

\section{Details of Testing Data}
\subsection{Degradations}

\label{sec:Basic Single Tasks}

We select 10 existing well-defined tasks as basic single tasks in the main paper.
They are \texttt{resize}, \texttt{blur}, \texttt{noise}, \texttt{compression}, \texttt{ringing}, \texttt{alg.artifact}, \texttt{damage}, \texttt{rain}, \texttt{haze} and \texttt{snow}.
We show a demo of these single degradations in \figurename~\ref{Fig:degradation_demo} (1 - 10), and put the degradation formulations and details as follow:
\paragraph{Resize.}
\begin{equation}
  I_{LQ}=Upsample(Downsample(I_{GT})),
\end{equation}
where $Downsample()$ is a common operation to generate low-resolution images in super-resolution tasks~\cite{SRCNN}, contains nearest-neighbor interpolation, bilinear interpolation, bicubic interpolation, and so on. 
In this work, to combine other degradations, we use $Upsample()$ to recover the low-resolution images to the original scale. We use the X4 scale bicubic as the $Downsample()$ and $Upsample()$ for the representative and random tasks.

\paragraph{Blur.}
\begin{equation}
  I_{LQ}=I_{GT}\circledast k,
\end{equation}
where $k$ is blur kernel. The isotropic and anisotropic Gaussian filters are common choices~\cite{SRMD}. 
We use the isotropic Gaussian kernel with a kernel size of 15, $\sigma = 2$ for the representative task. For random blurring tasks, we sample the kernel size uniformly from 7 to 23, and sample $\sigma$ uniformly from 0.2 to 3.

\paragraph{Noise.}
\begin{equation}
  I_{LQ}=I_{GT}+\epsilon,
\end{equation}
where the $\epsilon$ can be synthesized by Gaussian noise, Poisson noise~\cite{SIDD}, \etc. 
We use the Gaussian noise with a standard deviation of 20 for the representative task and sample standard deviation from 1 to 30 uniformly for random noise tasks.

\paragraph{Compression Artifacts.}
\begin{equation}
  I_{LQ}=Compress(I_{GT}).
\end{equation}
Lossy compression post-processing, \eg, JPEG and JPEG2000, will bring block artifacts. 
We use the JPEG with a compression quality of 50 for the representative task and sample compression quality from 30 to 95 uniformly for random tasks.

\paragraph{Ringing.}
\begin{equation}
  I_{LQ}=I_{GT}\circledast k,
\end{equation}
where $k$ is $sinc$ filter kernel. 
It is an idealized filter that cuts off high frequencies, to synthesize ringing~\cite{RealESRGAN}. 
We use the $sinc$ filter kernel with a kernel size of 15, and $\omega = 1.2$ for the representative task. We uniformly sample the kernel size from 7 to 23, and $\omega$ from $\pi$/3 to $\pi$ for random tasks.

\paragraph{Restoration Algorithm Artifacts.}
\begin{equation}
  I_{LQ}=Algorithm(I_{GT}).
\end{equation}
Many images from the Internet or smartphone cameras always have been processed by some simple restoration algorithm \eg Richardson-Lucy algorithm\cite{R-L1,R-L2}. 
These algorithms make images look better but also bring grain and smearing-effect artifacts. 
We regard these artifacts as a type of degradations. We use the Richardson-Lucy algorithm as the $Algorithm()$ for the representative and random tasks.

\paragraph{Damage.}
\begin{equation}
  I_{LQ}=\textbf{Damage}(I_{GT}).
\end{equation}
The goals of inpainting are various, from the restoration of damaged images to the removal/replacement of selected objects~\cite{bertalmio2000image}. 
In GIR, we only focus on low-level vision damages such as scratches and lines. 
We use the white or black lines with the number 10 and thickness of 7 for representative tasks and the number 5 to 10, and thickness of 5 to 10 for random tasks.

\paragraph{Rain.}
\begin{equation}
  I_{LQ}=I_{LQ}+rain(streak \& drop),
\end{equation}
where the $rain$ is usually generated by the appearance and imaging process of rain (most from photoshop)~\cite{Rain100,derainnet}.
There are works to collect real rain data or generate the rain streak or raindrop by GAN~\cite{wang2021rain}. 
We use the PhotoShop rain streaks synthesis method \footnote{This method is from https://www.photoshopessentials.com/photo-effects/photoshop-weather-effects-rain/} with strength = 75 for representative tasks and strength = 50 to 100 for random tasks.

\paragraph{Haze.}
\begin{equation}
  I_{LQ}=I_{GT} t(I_{GT}) + A (1 - t(I_{GT})),
\label{eq:haze}
\end{equation}
there are two critical parameters: $ A $ denotes the global atmospheric light, and $ t(I_{GT}) $ is the transmission matrix defined as:

\begin{equation}
t\left( I_{GT}\right) =e^{-\beta d\left( I_{GT}\right)},
\end{equation}
where $ \beta $ is the scattering coefficient of the atmosphere, and $ d\left( I_{GT}\right) $ is the distance between the object and the camera. 
Nowadays, to obtain enough haze and haze-free pairs, researchers usually synthesize haze images by the above formulations~\cite{RESIDE}. 
They first estimate the depth map and then sample the value of $ \beta $ and $A$ to generate haze images with different degrees.
We use the method of \cite{RESIDE} with  $ A $ = 0.9, $ \beta $ = 1.8 for representative tasks and $ A $ = 0.8 to 1, $ \beta $ = 0.5 to 2.5 for random tasks.

\paragraph{Snow.}

The degradation model of snow is under development. We select an acceptable model~\cite{chen2021all}:

\begin{equation}
  I_{LQ}=I_s t(I_{GT}) + A (1 - t(I_{GT})),
\end{equation}
where $ A $ denotes the global atmospheric light, and $ t(I_{GT}) $ is the transmission matrix to construct the veiling effect of snow. 
These two parameters are the same as that in haze generation (see Eq.~\ref{eq:haze}). $I_s$ is the snowy images without the veiling effect, which can be formulated as

\begin{equation}
  I_s=I_{GT}(1- Z R )+ C Z R,
\end{equation}

where $R$ is a binary mask which presents the snow location information, $C$ and $Z$ are the chromatic aberration map for snow images and the snow mask, respectively. 
We use the method of \cite{RESIDE} with  $ A $ = 0.9, $ \beta $ = 0.75 for representative tasks and $ A $ = 0.8 to 0.95, $ \beta $ = 0.5 to 1 for random tasks, the snow makes are following~\cite{chen2021all}.

\subsection{Selecting Method of Representative Tasks}
\label{sec:Selecting Method}
In the main paper, we select 50 representative tasks including single tasks.
The ten single tasks are described in Section~\ref{sec:Basic Single Tasks}.
For other tasks with the degradation order numbers $k$ are 2 to 5, we first generate lots of possible mixture tasks using random sequence and fixed strengths (See Section~\ref{sec:Basic Single Tasks}).
Then we execute clustering in each order following the method of\cite{ng2001spectral,von2007tutorial}.
Specifically, we use the histogram similarity to construct the similarity matrix by all image samples. 
After that, the similarity matrix is used as the input of spectral clustering, which can generate separable clusters in the GIR degradation space.
Finally, we utilize the degradation parameters of the cluster center as the representative tasks, which will be used to generate the test datasets for each order.
%
%
In that way, we obtain the test datasets of 50 representative tasks.

\subsection{Representative Tasks and Random Tasks}
\begin{figure*}[t]
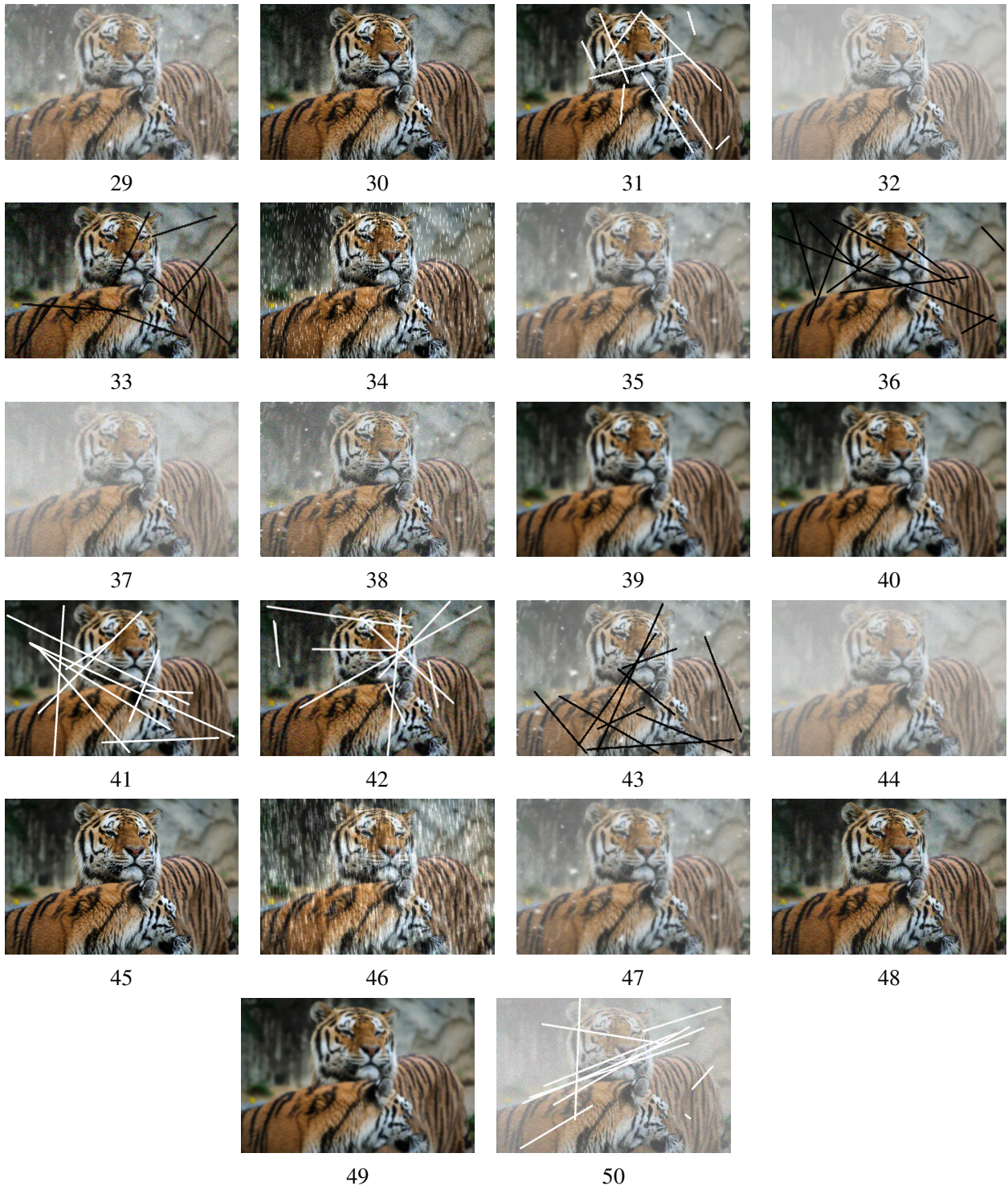


\scriptsize
\centering
\resizebox{1\textwidth}{!}{
%
    \end{adjustbox}
    \\

\end{tabular}}

\caption{The image demo of the basic single tasks (1 - 10), representative tasks (1 - 50). Zoom in for a better view.}
\label{Fig:degradation_demo}

\end{figure*}

As described in Section~\ref{sec:Selecting Method}, we select representative tasks of degradations with moderate strengths.
After that, we also generate 50 mixed tasks of degradations with random orders and strengths.
We show the degradation details and AR/ER performance on \tablename~\ref{table:tasks}, and the demo images on \figurename~\ref{Fig:degradation_demo}.
Note that, the detailed degradation orders and strengths of 50 random tasks (id 51 - 100) will not be released.
Because these tasks are cross-validation set, training a special MIR 100 that only concentrates on the 100 tasks will become possible once they are public.
It will hack this evaluation system, which is contrary to our original purpose.

\section{Details of Training Data}
We sample the clean ground truth images from DIV2K~\cite{DIV2K} and Flickr2K~\cite{timofte2017ntire} datasets, including 3,450 images in total.
We first estimate the depth of these images using the model of ~\cite{7346484}.
After that, we add the \texttt{rain}, \texttt{haze} and \texttt{snow} separately using the range of random degradation strengths in Section~\ref{sec:Basic Single Tasks}. 
Only the \texttt{haze} and \texttt{snow} need the depth information.
Finally, we obtain in total 13,800 images.
The remaining degradations are applied to image patches on the fly with the range of random degradation strengths in Section~\ref{sec:Basic Single Tasks}.

\section{Details of Experiments}


\subsection{Details of Model Performance}

We show the detailed results (See \tablename~\ref{table:baseline}) of the paragraph ``Model Generality'' in the main paper.
These models are far from the ideal GIR model.
Most models cannot maintain acceptable performance on most tasks (See the blue text).

\subsection{Details of Exploration}
\paragraph{Details of GIR vs MIR}
We show the detailed results (See \tablename~\ref{table:MIRvsGIR}) of the paragraph ``GIR vs MIR'' in the main paper.
It shows that although the MIR models can achieve acceptable even some excellence results on their training tasks, they can hardly generalize to other similar tasks.

\paragraph{Details of Training Data Scale Effect}
We show the detailed results (See \tablename~\ref{table:data}) of the paragraph ``Training Data Scale Effect'' in the main paper.
It proves that the core bottleneck of GIR is not dataset scale.
Surprisingly, adding ImageNet~\cite{ImageNet} images into the training set decreases the GIR performance.
This confirms our conclusion that GIR needs high-quality images for training and testing, while the image quality of ImageNet is not desirable, containing noise and compression.
Therefore, high-quality training data is still in demand for GIR.

\paragraph{Details of Training Strategy Effect}
We show the detailed results of the paragraph ``Training Strategy Effect'' in the main paper.
\tablename~\ref{table:ps} is the effect of training patch size and \tablename~\ref{table:bs} is the effect of training batch size.
we can see that all three metrics are improving with the increasing batch and patch size.
But the marginal effect of this improvement is diminishing
We also find that models with large batch and patch sizes may achieve the excellence line in some degradations.
All these degradations included \texttt{snow} or \texttt{rain}, illustrating that batch and patch sizes may have a greater impact on such degradations.


\begin{table*}[]
\centering
\begin{tabular}{|c|c|c|c|}
\hline
\rowcolor{color3} Task ID  & Degradations   & AR Model & ER Model          \\
\midrule

1   &  \texttt{resize}                                                                                                                                                      & 25.33    & 26.10    \\
2   & \texttt{Blur}                                                                                                                                                        & 30.74    & 32.41    \\
3   & \texttt{Noise}                                                                                                                                                       & 32.32    & 33.61    \\
4   & \texttt{compression}                                                                                                                                                 & 32.79    & 33.81    \\
5   & \texttt{damage}                                                                                                                                                      & 31.93    & 35.81    \\
6   & \texttt{ringing}                                                                                                                                                     & 29.91    & 35.03    \\
7   & \texttt{alg.artifact}                                                                                                                                                & 32.34    & 38.64    \\
\hline
\end{tabular}
\end{table*}

\begin{table*}[]
\centering
\begin{tabular}{|c|c|c|c|}
\hline
\rowcolor{color3} 
Task ID  & \qquad\qquad\qquad\qquad\quad\qquad Degradations   \qquad \enspace \qquad  \qquad \qquad                       \qquad\qquad                                                                                                                       & AR Model & ER Model  \\
\midrule

51  & -                                                                                                                                                                                     & 25.34    & 26.10    \\
52  & -                                                                                                                                                                                     & 32.12    & 33.76    \\
53  & -                                                                                                                                                                                     & 31.00    & 32.29    \\
54  & -                                                                                                                                                                                     & 31.65    & 32.65    \\
55  & -                                                                                                                                                                                     & 32.27    & 36.11    \\
56  & -                                                                                                                                                                                     & 32.49    & 37.47    \\
57  & -                                                                                                                                                                                     & 32.34    & 38.64    \\
\hline
\end{tabular}
\caption{Degradation details of the basic single tasks (1 - 10), representative tasks (1 - 50) and random tasks (51 - 100). We also show the PSNR (dB) of corresponding AR and ER models.}
\label{table:tasks}
\end{table*}

\begin{table*}[]
\centering
\begin{tabular}{|c|c|c|c|c|c|c|c|c|}
\hline
\rowcolor{color3} 
Task ID & AR Model & ER Model & DRL                          & RRDB                         & SwinIR                       & Uformer                      & Restormer                    & HAT                          \\ \midrule
1   & 25.33 & 26.10                        & 24.76                        & 24.98                        & 24.21                        & 24.66                        & 23.12                        & 24.58                        \\
2   & 30.74 & 32.41                        & 28.02                        & 28.56                        & 27.09                        & 27.45                        & 30.31                        & 27.74                        \\
3   & 32.32 & 33.61                        & 29.56                        & 31.38                        & 30.13                        & 30.28                        & 32.13                        & 30.58                        \\
4   & 32.79 & 33.81                        & 31.22                        & 31.44                        & 30.53                        & 30.35                        & 32.37                        & 31.26                        \\
5   & 31.93 & {\color[HTML]{0000FF} 35.81} & {\color[HTML]{0000FF} 31.95} & 31.92                        & {\color[HTML]{0000FF} 30.85} & {\color[HTML]{0000FF} 30.20} & {\color[HTML]{0000FF} 34.10} & {\color[HTML]{0000FF} 32.13} \\
6   & 29.91 & 35.03                        & 27.89                        & 27.92                        & 27.33                        & 27.70                        & 29.13                        & 27.88                        \\
7   & 32.34 & 38.64                        & 31.43                        & 32.31                        & 28.64                        & 28.58                        & 32.94                        & 29.50                        \\
\hline
\end{tabular}
\end{table*}

\begin{table*}[]
\centering
\begin{tabular}{|c|c|c|c|c|c|c|c|c|}
\hline\rowcolor{color3} 
Task ID & AR Model & ER Model & DRL                          & RRDB                         & SwinIR                       & Uformer                      & Restormer                    & HAT                          \\ \midrule

51  & 25.34 & 26.10                        & 24.76                        & 24.98                        & 24.21                        & 24.66                        & 23.12                        & 24.58                        \\
52  & 32.12 & 33.76                        & 28.93                        & 29.41                        & 27.54                        & 27.92                        & 31.31                        & 28.43                        \\
53  & 31.00 & 32.29                        & 27.91                        & 30.31                        & 29.13                        & 29.39                        & 30.93                        & 29.58                        \\
54  & 31.65 & 32.65                        & 30.37                        & 30.48                        & 29.59                        & 29.31                        & 31.32                        & 30.26                        \\
55  & 32.27 & 36.11                        & 31.94                        & 32.11                        & 31.05                        & {\color[HTML]{0000FF} 30.45} & {\color[HTML]{0000FF} 34.51} & {\color[HTML]{0000FF} 32.43} \\
56  & 32.49 & 37.47                        & 30.10                        & 30.09                        & 29.61                        & 29.61                        & 31.61                        & 30.25                        \\
57  & 32.34 & 38.64                        & 31.43                        & 32.31                        & 28.64                        & 28.58                        & 32.94                        & 29.50                        \\
\hline
\end{tabular}

\caption{Detailed PSNR (dB) of different GIR 
beseline models (paragraph ``Model Generality'' in the main paper). The results that surpass the corresponding AR/ER model are marked with {\color[HTML]{0000FF}Blur} and {\color[HTML]{FF0000}Red}. No model can surpass the corresponding ER models.}
\label{table:baseline}

\end{table*}

\begin{table*}[]
\centering
\begin{tabular}{|c|c|c|c|c|c|}
\hline\rowcolor{color3} 
Task ID & AR Model & ER Model & MIR 10                       & MIR 50                       & GIR                          \\ \midrule
1       & 25.33    & 26.10    & 25.11                        & {\color[HTML]{0000FF} 25.72} & 24.98                        \\
2       & 30.74    & 32.41    & 30.40                        & {\color[HTML]{0000FF} 31.68} & 28.56                        \\
3       & 32.32    & 33.61    & 30.16                        & {\color[HTML]{0000FF} 33.05} & 31.38                        \\
4       & 32.79    & 33.81    & 31.88                        & {\color[HTML]{0000FF} 33.14} & 31.44                        \\
5       & 31.93    & 35.81    & {\color[HTML]{0000FF} 32.78} & {\color[HTML]{0000FF} 34.01} & 31.92                        \\
6       & 29.91    & 35.03    & {\color[HTML]{0000FF} 31.74} & {\color[HTML]{0000FF} 34.12} & 27.92                        \\
7       & 32.34    & 38.64    & {\color[HTML]{0000FF} 34.25} & {\color[HTML]{0000FF} 37.54} & 32.31                        \\
\hline
\end{tabular}
\end{table*}

\begin{table*}[]
\centering
\begin{tabular}{|c|c|c|c|c|c|}
\hline\rowcolor{color3} 
Task ID & AR Model & ER Model & MIR 10                       & MIR 50                       & GIR                          \\ \midrule

51      & 25.34    & 26.10    & 25.11                        & {\color[HTML]{0000FF} 25.72} & 24.98                        \\
52      & 32.12    & 33.76    & 25.48                        & 23.95                        & 29.41                        \\
53      & 31.00    & 32.29    & 29.26                        & 30.49                        & 30.31                        \\
54      & 31.65    & 32.65    & 30.93                        & {\color[HTML]{0000FF} 31.89} & 30.48                        \\
55      & 32.27    & 36.11    & {\color[HTML]{0000FF} 32.89} & {\color[HTML]{0000FF} 34.32} & 32.11                        \\
56      & 32.49    & 37.47    & 24.72                        & 17.35                        & 30.09                        \\
57      & 32.34    & 38.64    & {\color[HTML]{0000FF} 34.25} & {\color[HTML]{0000FF} 37.54} & 32.31                        \\
\hline
\end{tabular}
\caption{Detailed PSNR (dB) of MIR and GIR models (paragraph ``GIR vs MIR'' in the main paper). The results that surpass the corresponding AR/ER model are marked with {\color[HTML]{0000FF}Blur} and {\color[HTML]{FF0000}Red}.}
\label{table:MIRvsGIR}
\end{table*}

\begin{table*}[]
\centering
\begin{tabular}{|c|c|c|c|c|c|c|}
\hline\rowcolor{color3} 
Task ID & AR Model & ER Model & Quarter of DF2K              & Half of DF2K                 & All of DF2K                  & ImageNet  +  DF2K            \\ \midrule
1       & 25.33    & 26.10    & 24.99                        & 24.85                        & 24.98                        & 23.04                        \\
2       & 30.74    & 32.41    & 28.53                        & 28.30                        & 28.56                        & 28.40                        \\
3       & 32.32    & 33.61    & 31.39                        & 31.22                        & 31.38                        & 31.30                        \\
4       & 32.79    & 33.81    & 31.28                        & 31.20                        & 31.44                        & 31.03                        \\
5       & 31.93    & 35.81    & {\color[HTML]{0000FF} 31.94} & 31.48                        & 31.92                        & 30.55                        \\
6       & 29.91    & 35.03    & 28.07                        & 27.57                        & 27.92                        & 28.01                        \\
7       & 32.34    & 38.64    & {\color[HTML]{0000FF} 32.40} & 32.25                        & 32.31                        & {\color[HTML]{0000FF} 33.45} \\
\hline
\end{tabular}
\end{table*}

\begin{table*}[]
\centering
\begin{tabular}{|c|c|c|c|c|c|c|}
\hline\rowcolor{color3} 
Task ID & AR Model & ER Model & Quarter of DF2K              & Half of DF2K                 & All of DF2K                  & ImageNet  +  DF2K            \\ \midrule

51      & 25.34    & 26.10    & 24.99                        & 24.85                        & 24.98                        & 23.04                        \\
52      & 32.12    & 33.76    & 29.37                        & 29.06                        & 29.41                        & 29.24                        \\
53      & 31.00    & 32.29    & 30.33                        & 30.16                        & 30.31                        & 30.27                        \\
54      & 31.65    & 32.65    & 30.36                        & 30.11                        & 30.48                        & 30.04                        \\
55      & 32.27    & 36.11    & 32.12                        & 31.61                        & 32.11                        & 30.72                        \\
56      & 32.49    & 37.47    & 30.27                        & 29.86                        & 30.09                        & 29.68                        \\
57      & 32.34    & 38.64    & {\color[HTML]{0000FF} 32.40} & 32.25                        & 32.31                        & {\color[HTML]{0000FF} 33.45} \\
\hline
\end{tabular}
\caption{Detailed PSNR (dB) of GIR-RRDB trained with different scale of data (paragraph ``Training Data Scale Effect'' in the main paper). The results that surpass the corresponding AR/ER model are marked with {\color[HTML]{0000FF}Blur} and {\color[HTML]{FF0000}Red}. No model can surpass the corresponding ER models.}
\label{table:data}
\end{table*}

\begin{table*}[]
\centering
\begin{tabular}{|c|c|c|c|c|c|c|}
\hline\rowcolor{color3} 
Task ID & AR Model & ER Model & PS 64                        & PS 128                       & PS 192                       & PS 256                       \\ \midrule
1       & 25.33    & 26.10    & 24.71                        & 24.98                        & 25.10                        & 25.25                        \\
2       & 30.74    & 32.41    & 27.85                        & 28.56                        & 28.83                        & 29.26                        \\
3       & 32.32    & 33.61    & 30.75                        & 31.38                        & 31.61                        & 31.71                        \\
4       & 32.79    & 33.81    & 30.79                        & 31.44                        & 31.60                        & 31.62                        \\
5       & 31.93    & 35.81    & 29.89                        & 31.92                        & {\color[HTML]{0000FF} 32.39} & {\color[HTML]{0000FF} 33.05} \\
6       & 29.91    & 35.03    & 27.52                        & 27.92                        & 28.02                        & 28.47                        \\
7       & 32.34    & 38.64    & 31.37                        & 32.31                        & {\color[HTML]{0000FF} 32.91} & {\color[HTML]{0000FF} 33.48} \\
\hline
\end{tabular}
\end{table*}

\begin{table*}[]
\centering
\begin{tabular}{|c|c|c|c|c|c|c|}
\hline\rowcolor{color3} 
Task ID & AR Model & ER Model & PS 64                        & PS 128                       & PS 192                       & PS 256                       \\ \midrule

51      & 25.34    & 26.10    & 24.71                        & 24.98                        & 25.10                        & 25.25                        \\
52      & 32.12    & 33.76    & 28.55                        & 29.41                        & 29.72                        & 30.16                        \\
53      & 31.00    & 32.29    & 29.69                        & 30.31                        & 30.49                        & 30.60                        \\
54      & 31.65    & 32.65    & 29.86                        & 30.48                        & 30.58                        & 30.60                        \\
55      & 32.27    & 36.11    & 30.08                        & 32.11                        & {\color[HTML]{0000FF} 32.56} & {\color[HTML]{0000FF} 33.25} \\
56      & 32.49    & 37.47    & 29.59                        & 30.09                        & 30.55                        & 30.81                        \\
57      & 32.34    & 38.64    & 31.37                        & 32.31                        & {\color[HTML]{0000FF} 32.91} & {\color[HTML]{0000FF} 33.48} \\
\hline
\end{tabular}
\caption{Detailed PSNR (dB) of GIR-RRDB trained with different patch sizes (paragraph ``Training
Strategy Effect'' in the main paper). The results that surpass the corresponding AR/ER model are marked with {\color[HTML]{0000FF}Blur} and {\color[HTML]{FF0000}Red}.}
\label{table:ps}
\end{table*}

\begin{table*}[]
\centering
\begin{tabular}{|c|c|c|c|c|c|c|}
\hline\rowcolor{color3} 
Task ID & AR Model & ER Model & BS4                          & BS8                          & BS16                         & BS32                         \\ \midrule
1       & 25.33    & 26.10    & 24.81                        & 24.98                        & 25.10                        & 25.33                        \\
2       & 30.74    & 32.41    & 28.11                        & 28.56                        & 28.93                        & 29.54                        \\
3       & 32.32    & 33.61    & 31.04                        & 31.38                        & 31.54                        & 31.92                        \\
4       & 32.79    & 33.81    & 30.98                        & 31.44                        & 31.49                        & 31.89                        \\
5       & 31.93    & 35.81    & 30.75                        & 31.92                        & {\color[HTML]{0000FF} 32.61} & {\color[HTML]{0000FF} 33.57} \\
6       & 29.91    & 35.03    & 27.55                        & 27.92                        & 28.20                        & 28.54                        \\
7       & 32.34    & 38.64    & 31.49                        & 32.31                        & {\color[HTML]{0000FF} 33.16} & {\color[HTML]{0000FF} 34.22} \\
\hline
\end{tabular}
\end{table*}

\begin{table*}[]
\centering
\begin{tabular}{|c|c|c|c|c|c|c|}
\hline\rowcolor{color3} 
Task ID & AR Model & ER Model & BS4                          & BS8                          & BS16                         & BS32                         \\ \midrule
51      & 25.34    & 26.10    & 24.81                        & 24.98                        & 25.10                        & 25.33                        \\
52      & 32.12    & 33.76    & 28.83                        & 29.41                        & 29.81                        & 30.47                        \\
53      & 31.00    & 32.29    & 29.97                        & 30.31                        & 30.50                        & 30.74                        \\
54      & 31.65    & 32.65    & 30.10                        & 30.48                        & 30.55                        & 30.77                        \\
55      & 32.27    & 36.11    & 30.91                        & 32.11                        & {\color[HTML]{0000FF} 32.84} & {\color[HTML]{0000FF} 33.83} \\
56      & 32.49    & 37.47    & 29.78                        & 30.09                        & 30.56                        & 31.24                        \\
57      & 32.34    & 38.64    & 31.49                        & 32.31                        & {\color[HTML]{0000FF} 33.16} & {\color[HTML]{0000FF} 34.22} \\
\hline
\end{tabular}
\caption{Detailed PSNR (dB) of GIR-RRDB trained with different batch sizes (paragraph ``Training
Strategy Effect'' in the main paper). The results that surpass the corresponding AR/ER model are marked with {\color[HTML]{0000FF}Blur} and {\color[HTML]{FF0000}Red}.}
\label{table:bs}
\end{table*}

\end{document}